\documentclass{article}

\usepackage[final,dblblindworkshop]{neurips_2026}
\workshoptitle{AIDaR: AI Data Readiness for Scientific Discovery}

\usepackage[utf8]{inputenc}
\usepackage[T1]{fontenc}
\usepackage{amsmath}
\usepackage{booktabs}
\usepackage{enumitem}
\usepackage{graphicx}
\usepackage{microtype}
\usepackage{placeins}
\usepackage{pdflscape}
\usepackage{tabularx}
\usepackage{xcolor}
\usepackage{xurl}

\graphicspath{{figures/}}
\setlist{nosep,leftmargin=*}

\newcommand{\dataset}[1]{\texttt{#1}}

\graphicspath{{figures/hsu_mcb_calibrated_full_grid_2026-08-29/}{figures/}}

\title{Calibrated Uncertainty Quantification on \mbox{TDC ADMET:}\\
What Can Be Resolved Across 22 Datasets?}
\author{%
  Laurits Fredsgaard \quad Michael Riis Andersen \quad Mikkel N. Schmidt \\
  Department of Applied Mathematics and Computer Science \\
  Technical University of Denmark \\
  \texttt{laula@dtu.dk, miri@dtu.dk, mnsc@dtu.dk}
}

\begin{document}

\maketitle

\begin{abstract}
Absorption, distribution, metabolism, excretion, and toxicity (ADMET) models often rely on small,
heterogeneous datasets, yet their predictions may determine which compounds proceed to experimental
testing. Useful evaluations must therefore assess both predictive performance and the quality of
reported probabilities or intervals. We ask which common uncertainty-quantification methods perform
best on the 22 Therapeutics Data Commons (TDC) ADMET datasets and what this benchmark can resolve. We
compare eleven calibrated pipelines using nested five-fold cross-validation across ten chemical
partitions and five model seeds. Ensemble Monte Carlo (MC) dropout has the best average rank for TDC
performance, Brier score, and 90\% interval score. For TDC performance, however, it is not separated
from the other ensemble pipelines; for Brier and interval score, LightGBM is also not ruled out as
best. LightGBM wins more individual datasets than any other method for TDC performance and Brier
score, but performs poorly on several endpoints. Smaller evaluations can identify a different
leader and often fail to recover the broader comparison. These results make LightGBM and an ensemble
neural model complementary baselines. Comparative claims should be checked across multiple chemical
partitions using statistical analyses matched to the question being asked.
\end{abstract}

\section{Introduction}
\label{sec:introduction}

Models of absorption, distribution, metabolism, excretion, and toxicity (ADMET) help prioritize
compounds for synthesis and experimental testing. A useful model must therefore do more than make
accurate predictions: it should also indicate when they
are uncertain. Reliable uncertainty estimates can help decide which compounds to test, where more
data are needed, and when a prediction should not drive an experimental decision
\citep{hirschfeld2020uncertainty}.

ADMET datasets are often small, and a more complicated model is not necessarily better. Fingerprint
and descriptor models remain competitive across many small molecular datasets
\citep{notwell2023maplight,praski2025benchmarking}, while predictability differs substantially
across assays \citep{aleksic2022predictability}. Uncertainty quantification
(UQ) methods may likewise perform differently from one endpoint to another.

The comparison itself also varies. Chemical partitions, model initialization, and the data used for
tuning can change the observed differences between both point predictors and UQ methods
\citep{bouthillier2021accounting,ash2025practically}. This matters for the 22 public ADMET tasks in
the Therapeutics Data Commons (TDC), which use endpoint-specific performance measures and a fixed scaffold
test split \citep{huang2021tdc}. After our structure curation, the datasets contain 475 to 13,094
compounds. An audit of leading
submissions found reproducibility failures, data leakage, and model choices influenced by the public
test sets \citep{koleiev2026enduser}. We therefore ask which methods perform consistently across
datasets, which work best on particular endpoints, and whether those conclusions survive changes in
the evaluation design.

We study eleven calibrated pipelines based on LightGBM and single or ensemble neural models on all
22 datasets, crossing ten chemical partitions, five model seeds, and five outer folds. This grid
supports comparisons across datasets and on each dataset's original score scale, followed by
smaller-design stability analyses.

Our contributions are:
\begin{itemize}
  \item We compare eleven calibrated UQ pipelines on all 22 TDC ADMET datasets using nested
  cross-validation across multiple chemical partitions and model seeds.
  \item We show that ensemble Monte Carlo (MC) dropout is the most consistent method across datasets, while
  LightGBM wins many individual endpoints and performs poorly on others.
  \item We examine how reduced split and tuning designs affect both the apparent leader and the
  differences supported by the statistical analysis.
\end{itemize}

\section{Related work}
\label{sec:related}

\paragraph{Public ADMET benchmarks and data regimes.}
TDC brought established ADMET endpoints into a widely used collection with fixed scaffold splits
\citep{huang2021tdc}. A later audit of leading submissions found reproducibility, leakage, and
test-set reuse problems \citep{koleiev2026enduser}. Since TDC was introduced, newer resources have
extended ADMET evaluation through stronger curation, blinded or prospective tests, and multitask
structure. PharmaBench adds curation and experimental-condition extraction
\citep{niu2024pharmabench}, while the ASAP--Polaris--OpenADMET challenge used blinded, temporally
split measurements from an active lead-optimization programme
\citep{macdermottopeskin2025blind}. The Biogen and ExpansionRx releases provide industry assay
collections \citep{fang2023prospective,openadmet2026expansionrx}. These and public assay data also
support multitask studies: \citet{adrian2025multitask} fine-tune models on Biogen and public assay
data, while \citet{xue2026probabilistic} evaluate probabilistic pretraining on Biogen, ExpansionRx,
and ChEMBL-MT.

\paragraph{Simple molecular representations as baselines.}
Recent molecular benchmarks increasingly include graph neural networks and pretrained molecular
representations \citep{li2024muben,praski2025benchmarking}. Their gains over simpler representations
are not uniform. ADMET predictability can vary more by assay than by algorithm or data volume
\citep{aleksic2022predictability}, and MapLight obtained strong TDC results by combining fingerprints
and physicochemical descriptors with boosted trees \citep{notwell2023maplight}. In a separate study
across 25 classification datasets, most frozen pretrained molecular embeddings offered little or no
improvement over an extended-connectivity fingerprint baseline \citep{praski2025benchmarking}. A carefully tuned
fingerprint model is therefore an important reference for small molecular datasets.

\paragraph{What molecular UQ evaluations measure.}
The relevant UQ evaluation depends on how the prediction will be used. Ranking-based evaluation is
appropriate when the aim is to prioritize predictions or reject the least reliable ones. Decisions based on a
numerical probability or interval instead require uncertainty on a meaningful scale. UNIQUE makes
this distinction explicit by separating ranking-based, calibration-based, and proper-score
evaluations \citep{lanini2024unique}. Proper scores reward calibration together with sharpness;
Brier score and interval score are
examples for probabilities and intervals \citep{gneiting2007strictly}. Post-hoc transformations can
also change the ordering of models, making the choice of calibration procedure part of the
comparison design \citep{ranjan2024posthoc}. MUBen broadened molecular UQ comparisons across
representations and both classification and regression datasets \citep{li2024muben}. Other work has
tested whether error-ranking performance persists under chemical and temporal shifts
\citep{parrondopizarro2026shifts}. Results vary across tasks and measures
\citep{hirschfeld2020uncertainty}, while ensembles have provided strong results in several molecular
UQ benchmarks \citep{scalia2020evaluating,li2024muben}.

\paragraph{Chemical splits and method comparison.}
Random splits tend to preserve close chemical analogues across training and test sets. Scaffold or
similarity-based splits reduce structural overlap, while temporal splits can approximate prospective
use when reliable dates are available \citep{yang2019analyzing,sheridan2013timesplit}. Split,
optimization, and other stochastic choices all contribute to benchmark variation
\citep{bouthillier2021accounting,ash2025practically}. \citet{ash2025practically} recommend shared
resamples and paired comparisons, generally using five-by-five repeated cross-validation (CV) for
500--100,000 molecules and five-by-two CV below 500. They allow a single split for very large
datasets when repeated CV is computationally impractical. Their lower-cost
hyperparameter protocol uses one training--validation split inside each repeated-CV training set
rather than full inner CV.

The statistical procedure should follow the comparison being made. Across datasets, the Friedman
test asks whether method ranks differ overall \citep{demsar2006statistical}. Nemenyi
critical-difference diagrams provide a conventional all-pairs summary, while Holm-adjusted pairwise
tests can give more power and adjusted $p$-values for the chosen comparison family
\citep{garcia2008extension}. Within one dataset, Tukey's honestly significant difference procedure
is useful when every pair of method means is of interest \citep{ash2025practically}. Hsu's
multiple-comparisons-with-the-best (MCB) procedure focuses instead on which methods can be ruled out
as best \citep{hsu1984constrained}. Cross-dataset ranks address consistency across tasks; comparisons
on the original score scale address what can be distinguished on one task.

\section{Benchmark and evaluation}
\label{sec:methods}

\paragraph{Datasets and chemical partitions.}
We use all 22 TDC ADMET datasets as separate tasks: 13 binary and nine regression. Molecules are
standardized to parent structures with the ChEMBL pipeline \citep{bento2020chembl} and keyed by
parent Standard InChIKey. We keep repeated measurements but assign all records for one compound to
the same fold. Inputs concatenate 2048-dimensional radius-2 Morgan count fingerprints
\citep{rogers2010morgan} and 200 cumulative-distribution-normalized RDKit2D descriptors computed
with DescriptaStorus 2.8.0 \citep{kelley2024descriptastorus}. Every pipeline is
fitted on the log target for four endpoints:
\dataset{vdss\_lombardo}, \dataset{half\_life\_obach},
\dataset{clearance\_microsome\_az}, and \dataset{clearance\_hepatocyte\_az}. Predictions are
transformed back before scoring.

Five outer folds are formed from Taylor--Butina clusters at Tanimoto distance cutoff 0.5
\citep{butina1999clustering}. For each outer fold, we recompute the clusters within the outer-training
compounds and divide them into a dedicated calibration set and a model-development set. Complete
clusters are added to the calibration set until it contains approximately 20\% of the observations.
Only the model-development set enters five-fold inner cross-validation. We average each
configuration's five TDC-score trajectories step by step, choose its best step on that mean curve,
and select the configuration with the best resulting score. We refit it on the full model-development
set and fit the post-hoc calibrator on the untouched calibration set.

\paragraph{Prediction pipelines.}
We compare LightGBM \citep{ke2017lightgbm} with five neural pipelines: a deterministic multilayer
perceptron (MLP), MC dropout \citep{gal2016dropout}, checkpoint-resumed stochastic
weight averaging Gaussian (SWAG) \citep{maddox2019swag}, last-layer Laplace
\citep{daxberger2021laplace}, and a post-hoc variational Bayesian last layer (VBLL)
\citep{harrison2024vbll}. Their ten-member
counterparts are deep ensemble \citep{lakshminarayanan2017deep}, ensemble MC dropout
\citep{filos2019systematic}, MultiSWAG \citep{wilson2020bayesian}, mixture of Laplace approximations
(MoLA) \citep{eschenhagen2021mixtures}, and ensemble post-hoc VBLL, giving eleven pipelines. All ten
neural pipelines use the same MLP architecture. LightGBM is the only non-neural model, and every
pipeline uses the same molecular features.

LightGBM and the MLP backbones use binary cross-entropy for the 13 classification tasks, L1 loss for
the five regression tasks scored by mean absolute error, and L2 loss for the four scored by
Spearman correlation. Post-hoc VBLL minimizes the negative evidence lower bound (ELBO).

LightGBM and the deterministic MLP are selected from 32 presampled Sobol configurations using the
mean inner-fold TDC metric. All neural UQ pipelines inherit the MLP architecture, hyperparameters,
and training epoch selected for deterministic point prediction. MC dropout, SWAG, and Laplace use
the selected checkpoint; post-hoc VBLL fits a variational last layer over its frozen features. For
each feature map, we select among 81 last layers spanning nine prior and nine Wishart scales by the
final ELBO on the model-development rows; ensemble post-hoc VBLL repeats this for every member.
Ensembles combine ten independently initialized refits of the same selection. This holds backbone
selection fixed across neural UQ constructions. Fits run 2,000 LightGBM rounds or 300 MLP epochs,
with the deployed step selected from the inner validation trajectory.

Deterministic MLP and MC dropout use the same selected network and training-time dropout. At
inference, deterministic MLP disables dropout and uses one forward pass, whereas MC dropout keeps
dropout active and averages 30 stochastic passes. Starting from the selected checkpoint, SWAG runs
ten epochs of stochastic gradient descent with momentum and collects up to four parameter snapshots
per epoch from the first epoch. The resulting Gaussian approximation has diagonal-plus-rank-20 covariance
and is evaluated with 30 posterior weight draws. MultiSWAG
uses ten independently refitted SWAG members with equal weight, each evaluated with 30 posterior
weight draws. For regression, its combined variance includes both the average uncertainty within
members and the variance between member means. For binary prediction, we average the ten member
probabilities after averaging each member's 30 draws.

\paragraph{Calibration and evaluation measures.}
We estimate post-hoc calibration parameters from the dedicated calibration set. For binary tasks,
we apply temperature scaling to the logit of each pipeline's final marginal probability after
averaging ensemble members \citep{guo2017calibration}. We evaluate the resulting probabilities with
Brier score, a proper score that reflects both calibration and refinement.

Regression requires different calibration procedures because the pipelines supply different forms
of uncertainty. For LightGBM and the deterministic MLP, we estimate one constant residual standard
deviation. For MC dropout, SWAG, deep ensemble, ensemble MC dropout, and MultiSWAG, we estimate an
additional observation-noise variance and add it to the supplied predictive variance. For Laplace,
post-hoc VBLL, MoLA, and ensemble post-hoc VBLL, we multiply the predictive standard deviation by a
fitted non-negative factor. These adjusted distributions define central 90\% intervals, evaluated
by interval score. For the four log-trained endpoints, we estimate calibration parameters and form
intervals on the log target, then exponentiate the interval limits and score them in the endpoint's
original units.

Our main measures are the endpoint-specific TDC metric, Brier score for binary tasks, and interval
score for regression tasks. Repeated labels and predictions are averaged within compound for the
TDC metrics; Brier and interval scores use individual observations. The appendix also reports area
under the receiver operating characteristic curve, average precision, negative log-likelihood, and
expected calibration error for binary tasks, and mean absolute error, root mean squared error,
Spearman correlation, Gaussian negative log-likelihood, and coverage error for regression.

\paragraph{Full grid and comparison across datasets.}
For each dataset and method under each evaluation measure, we average exactly
$10\times5\times5=250$ outer-fold scores: ten separately generated chemical partitions, five model seeds,
and five outer folds. These 250 values estimate one score. We rank the eleven methods within each
dataset, with rank 1 best and average ranks for exact score ties. Each dataset contributes one score
and one rank to the cross-dataset comparison. The 1,826,000 from-scratch model fits are itemized in
Appendix~Table~\ref{tab:compute-counts}.

For each evaluation measure, the tie-corrected Friedman statistic with the Iman--Davenport $F$
approximation tests whether method ranks differ overall across datasets \citep{demsar2006statistical}.
We then compare all 55 method pairs using two-sided normal tests of their mean-rank differences and
apply Holm's procedure to this fixed all-pairs family, controlling the family-wise error rate at 5\%
\citep{garcia2008extension}. We report the comparisons involving the method with the best average
rank. A non-significant comparison means that the data do not rule out the alternative as best; it
does not establish equal performance.

\paragraph{Per-dataset comparisons.}
Average ranks do not show what can be resolved on one endpoint, so we also compare methods on each
dataset's original score scale. We average the five model seeds and five outer folds within each
chemical partition, giving 10 paired complete-partition scores per method. A repeated-measures
analysis of variance first tests whether any method differs; if so, Hsu's MCB procedure at the 5\%
level assesses which methods can be ruled out as best \citep{hsu1984constrained}. We call the
observed mean in the favorable direction the empirical best. Appendix~\ref{app:dataset-scores}
gives the decision threshold and its coverage interpretation.

The ten partitions are generated separately but reuse the same finite set of compounds; their
variation therefore measures sensitivity to the chemical partition rather than population-sampling
uncertainty for new compounds. The analysis also assumes normally distributed errors and one residual
variance across methods. Its $p$-values should therefore be read as conditional on these assumptions,
the observed datasets, and the completed evaluation grid.

\paragraph{Stability under smaller evaluation designs.}
We use finite-grid subsampling to ask how much of the completed evaluation is needed to recover its
conclusions. We compare one held-out fold with one or more complete five-fold CVs, using either one
model seed or the average of all five. Partition and fold levels are selected without replacement
and shared across methods and datasets. Every reported design enumerates all distinct paired
subsets. For each, we recalculate
the empirical winners, mean ranks, Holm decisions, and, when at least two blocks are available, the
per-dataset MCB conclusions. The resulting rates measure agreement with the completed grid.

\section{Results}
\label{sec:results}

\begin{figure*}[t]
  \centering
  \includegraphics[width=\textwidth]{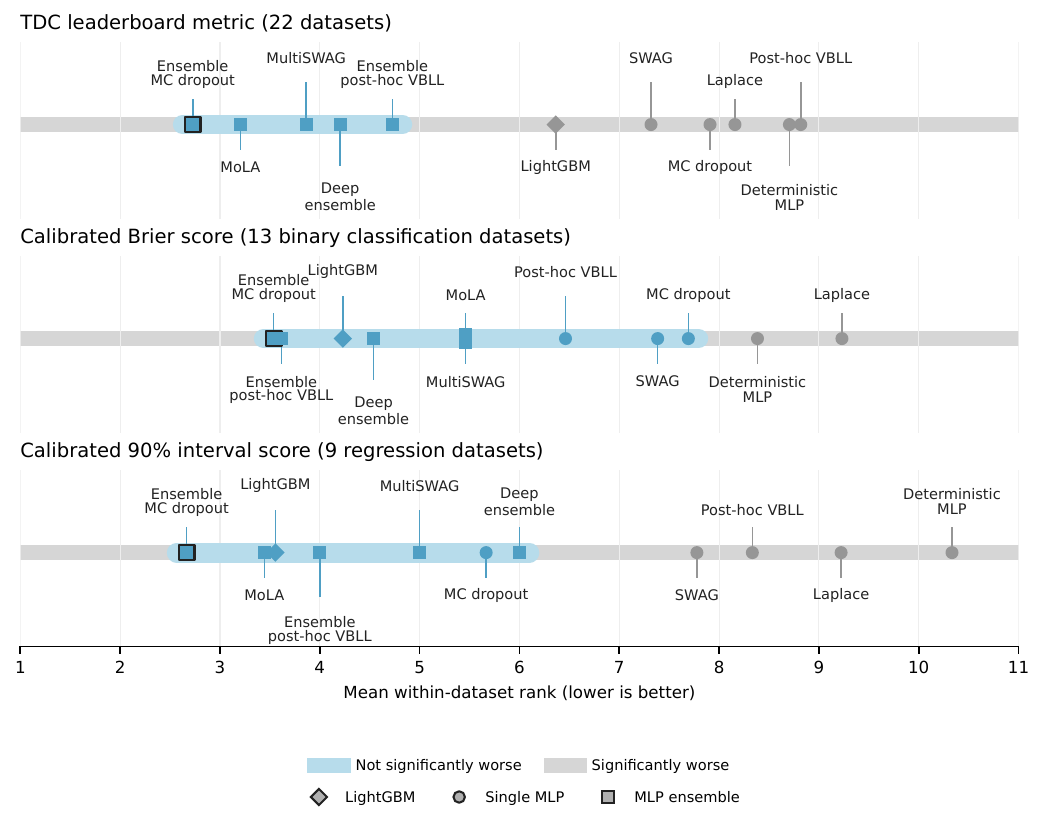}
  \caption{Mean ranks across datasets, with rank 1 best and the empirical leader outlined. Marker
  shape distinguishes LightGBM, single MLP, and MLP ensemble pipelines. ``Not significantly
  worse'' marks methods that the Holm-adjusted comparison does not distinguish from the leader;
  ``Significantly worse'' marks methods that it does. Adjustment is performed separately for each
  evaluation measure across all 55 method pairs.}
  \label{fig:global-rank-holm}
\end{figure*}

Across datasets, ensemble MC dropout has the best average rank for the TDC metric, calibrated Brier
score, and calibrated interval score (Figure~\ref{fig:global-rank-holm}). The Friedman tests find
clear differences among the methods for all three measures. The follow-up comparisons qualify the
ranking: the group not separated from ensemble MC dropout contains all five ensemble pipelines for TDC performance.
For Brier score, only the deterministic MLP and Laplace are separated from the leader. For interval
score, the group not separated from the leader includes LightGBM, MC dropout, and all five ensemble
pipelines.

Appendix~Table~\ref{tab:global-rank-holm} gives every mean rank and adjusted $p$-value.

LightGBM has the most empirical wins for both the TDC metric and Brier score, yet its performance is
strongly endpoint dependent. Its TDC ranks range from first to last, producing a substantially worse
average rank than ensemble MC dropout. For interval score, LightGBM and ensemble post-hoc VBLL each
win three datasets, while ensemble MC dropout ranks between first and fourth on every regression
dataset.

\begin{figure*}[t]
  \centering
  \includegraphics[width=\textwidth]{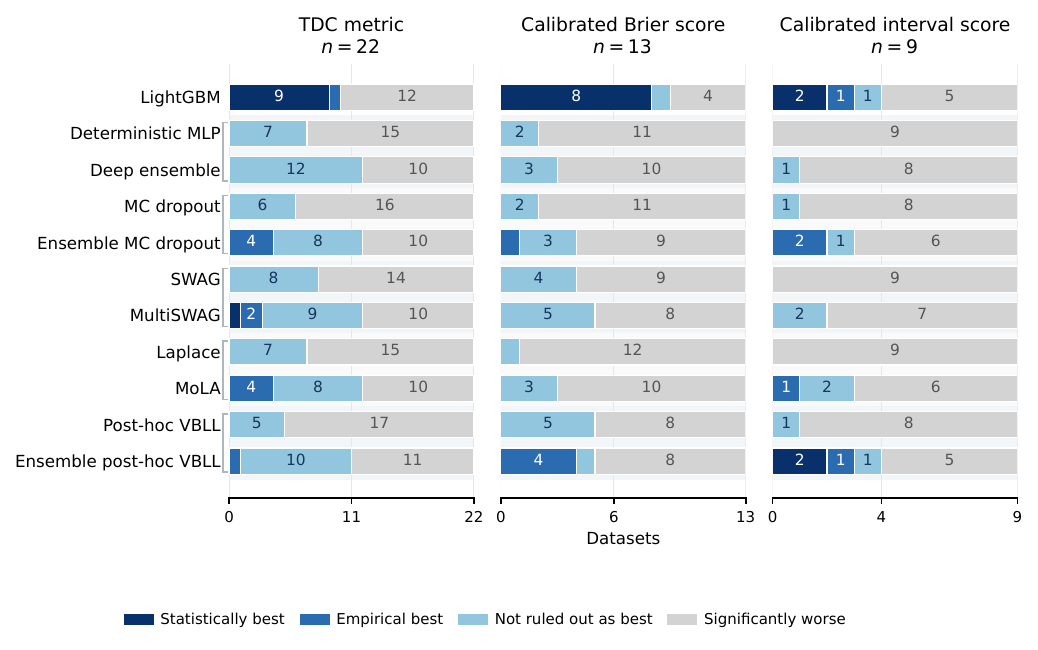}
  \caption{Per-dataset classifications from Hsu's MCB procedure after averaging the five model
  seeds and five outer folds within each of ten chemical partitions, giving ten paired
  complete-partition blocks. ``Statistically best'' means that the empirical winner is the only
  method not ruled out by MCB. ``Empirical best'' marks other observed winners, and ``Not ruled
  out as best'' marks additional methods not excluded by the test. The initial test finds a method
  difference for all 44 dataset-score combinations; each is assessed separately
  at the 5\% level, with the coverage interpretation given in
  Appendix~\ref{app:dataset-scores}.}
  \label{fig:per-dataset-status-tally}
\end{figure*}

\paragraph{What individual datasets can distinguish.}
Average ranks summarize consistency across the suite; comparisons on the original score scale show
which methods the repeated evaluations can distinguish on each endpoint. Figure~\ref{fig:per-dataset-status-tally}
makes the contrast clear. LightGBM is often the sole method not ruled out as best, but it is also
ruled out on many other endpoints. Ensemble MC dropout wins fewer datasets but is not ruled out as
best across a broader range of them. The evaluations find some method difference for all 44
dataset-score combinations.

\FloatBarrier

Appendix~\ref{app:dataset-scores} shows the corresponding scores and decisions for every dataset.

\begin{figure*}[p]
  \centering
  \includegraphics[width=\textwidth,height=0.82\textheight,keepaspectratio]
    {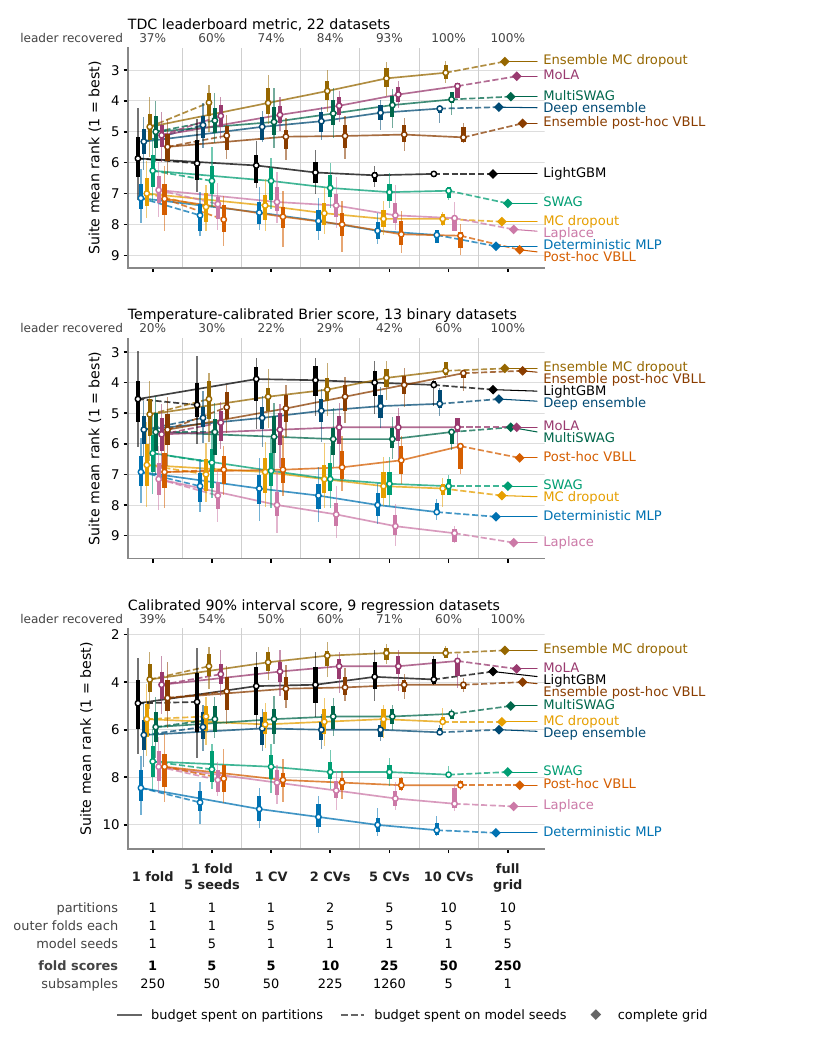}
  \caption{Suite-wide mean ranks under smaller evaluation designs. Open circles and bars show the
  median, interquartile range, and 5th--95th percentile range over paired finite-grid subsets;
  diamonds show the completed grid. Solid branches add folds and partitions, while dashed branches
  average model seeds. Percentages give the subsets that recover the sole full-grid leader. These
  empirical ranges are not confidence intervals.}
  \label{fig:rank-trajectories}
\end{figure*}

\paragraph{Evaluation-design stability.}
With one model seed, a single held-out fold selects the same leader as the completed grid in only
37.2\% of TDC-metric, 20.4\% of Brier, and 39.2\% of interval-score subsets
(Figure~\ref{fig:rank-trajectories}). The apparent leader from one fold may therefore differ from the
leader obtained after averaging the full evaluation. At the equal cost of five scores, one complete
cross-validation reproduces the full-grid rank ordering more closely than averaging five model seeds
on one held-out fold. Repeated complete cross-validations narrow the rank variation, with Brier score
settling most slowly.

\paragraph{Tuning with a single validation split.}
To test whether a single train--validation split is sufficient for tuning, we compare its selected
configuration with the choice from all five inner folds. Median agreement is 19.2\% for LightGBM and
17.8\% for the MLP across datasets, and is still incomplete when four folds are used. The one-fold
choices also score worse when evaluated across all five validation folds
(Appendix~Table~\ref{tab:inner-selection-stability}).
\FloatBarrier

\section{Discussion}
\label{sec:discussion}

\paragraph{Baseline choice.}
Ensemble MC dropout provides a consistently strong neural reference, but its best average rank does
not establish an advantage over the other ensemble pipelines. LightGBM's many individual wins and
poor results elsewhere show why consistency across a benchmark and performance on a particular
assay can differ. The per-dataset comparisons identify where one method has a clear advantage and
where several alternatives cannot be ruled out as best. A comparison using only one baseline family
can therefore give an incomplete picture of a new method's performance. Both LightGBM and an
ensemble neural baseline should be included when evaluating new molecular predictors, including
pretrained models.

\paragraph{Variation across evaluations.}
Evaluation introduces uncertainty of its own. Changing which compounds enter training and testing
can change the apparent leader, and averaging model seeds on one fold leaves variation between
chemical partitions unexplored. At the same number of fold-score evaluations, completing
cross-validation gives a more stable full rank ordering than repeating seeds on one fold. A compact
evaluation can help screen methods; a claim of improvement should be checked across multiple
chemical partitions. Paired comparisons account for the variation in method differences, while
multiplicity adjustment limits spurious claims of a difference when several methods are compared.

Shared training compounds weaken the independence assumption behind the per-dataset tests. Some
endpoint families also share compounds, and only nine datasets contribute to the cross-dataset
interval-score comparison. Its unresolved differences deserve caution: the absence of a detected
difference gives little reason to prefer one method over another on statistical grounds alone.

\paragraph{Tuning and computational cost.}
Validation data affect the comparison before test scores are calculated. A single validation split
often selects a different configuration from five-fold selection, and those choices perform worse
across the five validation folds. Authors proposing a new model should give simple baselines a fair
chance through careful hyperparameter and stopping-point selection. Their competitiveness here makes
that tuning part of establishing an improvement, including for expensive pretrained architectures.

Ensembles require additional model fits, while other UQ methods add posterior sampling or last-layer
optimization. We made the full benchmark feasible by sharing MLP hyperparameter and checkpoint
selection across UQ methods. These choices optimize point prediction. Selecting the final
ensemble or using a calibrated validation score could favor different hyperparameters or stopping
epochs \citep{ranjan2024posthoc,fredsgaard2025joint}. The rankings consequently describe the complete
pipelines evaluated here, including their selection procedures and additional computation.

\paragraph{Using the uncertainty estimates.}
The relevant score depends on what the predictions will be used for. Brier score assesses
probabilities through both calibration and refinement without separating epistemic and aleatoric
uncertainty. The regression results depend on the calibration maps and Gaussian assumptions used to
construct the intervals. An application that uses uncertainty mainly to rank candidates would also
need to assess that ranking.

Claims of improvement should be supported by comparisons with both carefully tuned baselines.
Reporting the size of the gain and its variation across chemical partitions helps readers judge
its practical value.

\nopagebreak[4]

\paragraph{Code and data availability.}
Code is available at \url{https://github.com/lauritsf/admet-uq}; data are available from DTU Data
\citep{larsen2026admetuqdata}.

\nopagebreak[4]
\paragraph{Acknowledgements.}
The authors acknowledge support from the Novo Nordisk Foundation under grant no.\
NNF22OC0076658 (Bayesian neural networks for molecular discovery). We acknowledge the Danish
e-infrastructure Consortium (DeiC) for awarding this project access to the LUMI supercomputer,
owned by the EuroHPC Joint Undertaking, hosted by CSC (Finland) and the LUMI consortium.

\nopagebreak[4]
\paragraph{Disclosure.}
Large language models assisted with prose refinement and code writing; the authors reviewed all
resulting material.

\clearpage
\bibliographystyle{plainnat}
\bibliography{references}

\clearpage
\appendix
\section{Calibration procedures and results}
\label{app:calibration-details}

Every calibrator is fitted on the dedicated calibration holdout inside the corresponding outer
training split. Table~\ref{tab:calibration-map} summarizes how each form of uncertainty is
calibrated. Figures~\ref{fig:binary-dataset-scores}, \ref{fig:appendix-native-binary-nll}, and
\ref{fig:appendix-native-binary-ece} show Brier score, binary negative log-likelihood (NLL), and
expected calibration error (ECE) by dataset. Figures
\ref{fig:regression-dataset-scores}, \ref{fig:appendix-native-regression-nll}, and
\ref{fig:appendix-native-regression-coverage-error} show interval score, Gaussian NLL, and coverage
error for regression.

\begin{table*}[!htbp]
\centering
\caption{Calibration applied to each pipeline. A supplied scale is a standard deviation on the
target used to train the model.}
\label{tab:calibration-map}
\footnotesize
\begin{tabularx}{\textwidth}{@{}>{\raggedright\arraybackslash}p{0.29\textwidth}
  >{\raggedright\arraybackslash}p{0.31\textwidth}>{\raggedright\arraybackslash}X@{}}
\toprule
Pipelines & Supplied uncertainty & Calibration applied \\
\midrule
All eleven binary pipelines & Marginal event probability & One scalar temperature \\
LightGBM; deterministic MLP & Point prediction only & Constant Gaussian residual scale \\
MC dropout; SWAG; deep ensemble; ensemble MC dropout; MultiSWAG
  & Stochastic-pass, posterior-draw, or ensemble-member spread
  & Additive observation-noise variance \\
Laplace; post-hoc VBLL; MoLA; ensemble post-hoc VBLL
  & Total predictive standard deviation & Standard-deviation multiplier \\
\bottomrule
\end{tabularx}
\end{table*}

For binary predictions, we choose the temperature that minimizes NLL on the calibration set and
divide the probability logits by this value. If the calibration labels cannot support a finite
temperature estimate, we set the temperature to one and keep the original probabilities.

For regression pipelines with only a point prediction, the fitted constant scale is the
root-mean-square calibration residual. For pipelines with a predictive standard deviation, the
scale multiplier is the root-mean-square residual after division by that standard deviation. For
pipelines supplying epistemic spread, we choose a non-negative observation-noise variance by
minimizing Gaussian NLL on the calibration set and add it to the supplied variance. Central 90\%
intervals extend 1.645 adjusted standard deviations on either side of the mean.

The four log-trained endpoints are calibrated on the log target. Exponentiating the interval limits
produces asymmetric intervals in the original units used for scoring. The appendix NLL values for
these endpoints use the corresponding log-normal density in the original units.

\begin{figure*}[p]
  \centering
  \includegraphics[width=\textwidth]{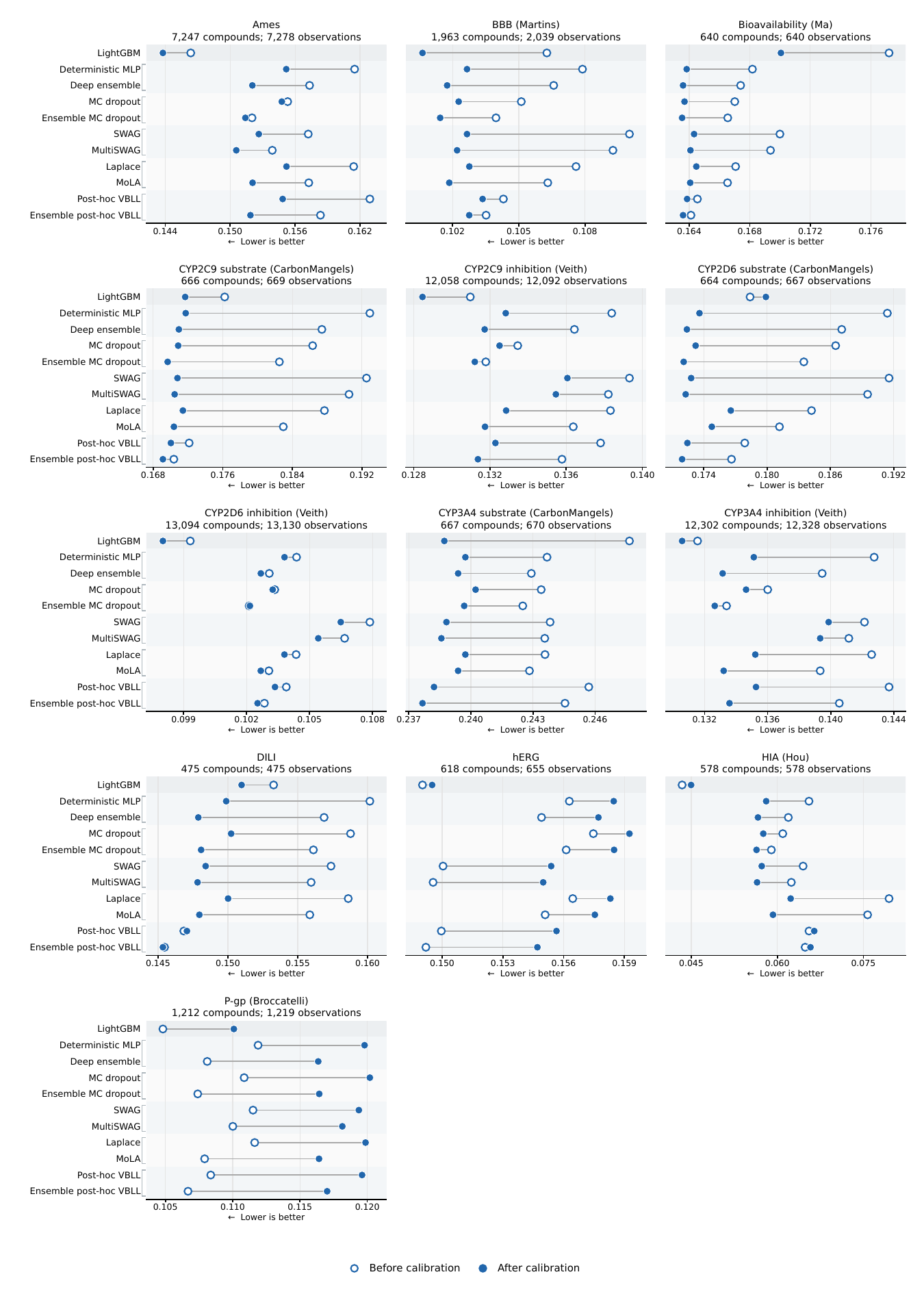}
  \caption{Full-grid mean Brier scores before and after temperature scaling for all 13 binary
  datasets and eleven pipelines. Open markers show uncalibrated scores and filled markers show
  calibrated scores. Lower is better.}
  \label{fig:calibration-brier-endpoints}
\end{figure*}

\begin{figure*}[p]
  \centering
  \includegraphics[width=\textwidth]{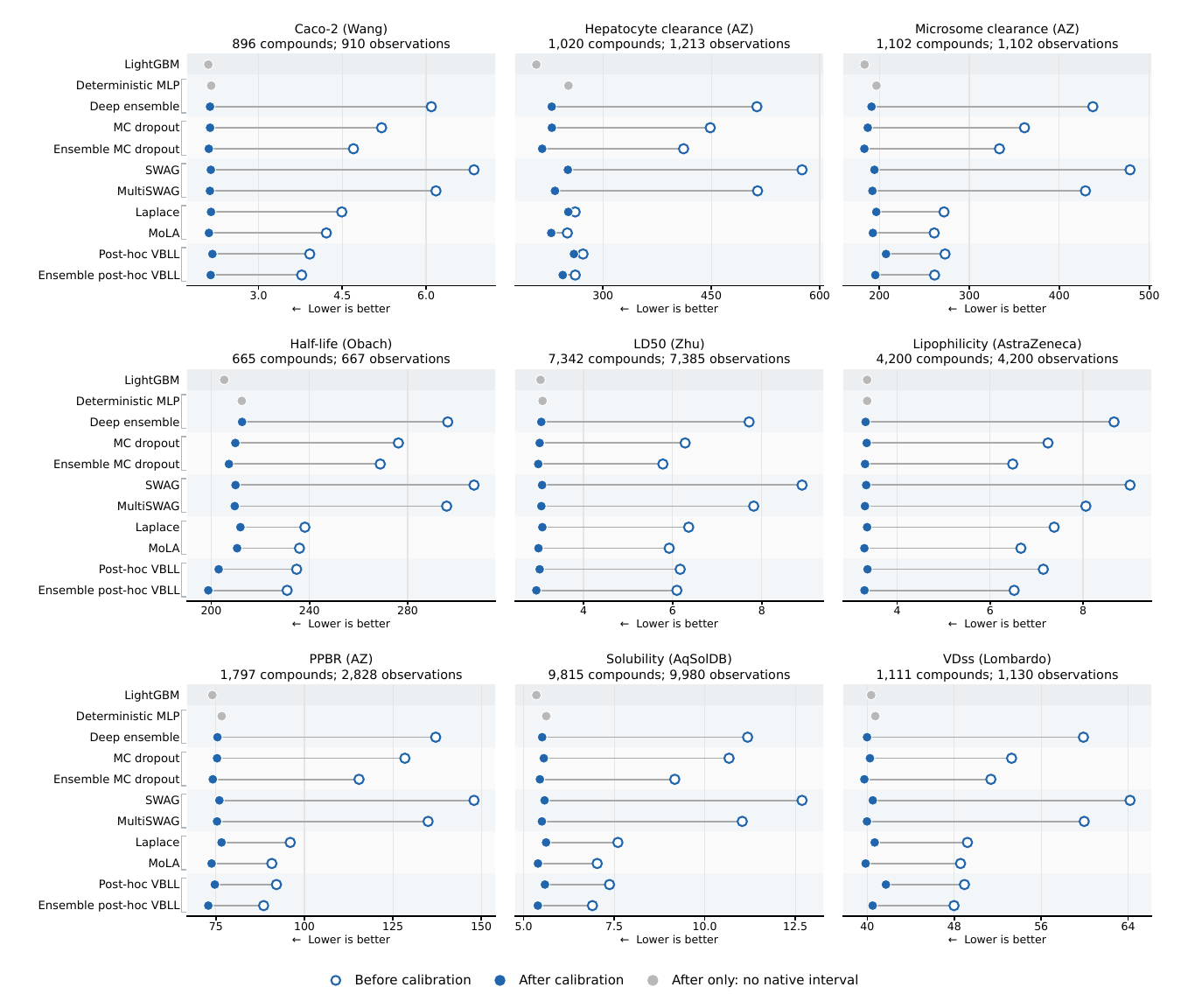}
  \caption{Full-grid mean 90\% interval scores before and after method-specific calibration for
  all nine regression datasets. Open markers show native intervals and filled markers show
  calibrated intervals. LightGBM and deterministic MLP supply no native interval, so only their
  calibrated scores are shown. Axes are endpoint-specific; lower is better.}
  \label{fig:calibration-interval-endpoints}
\end{figure*}

Figures~\ref{fig:calibration-brier-endpoints} and
\ref{fig:calibration-interval-endpoints} show the full-grid mean score for each method and endpoint
before and after calibration. Each point averages 250 outer-fold evaluations. LightGBM and the
deterministic MLP supply only point predictions for regression, so their interval-score panels begin
after the constant residual scale is fitted.

\section{Global rank comparisons}
\label{app:global-rank-details}

The Friedman test finds differences among methods for all three evaluation measures
(Table~\ref{tab:global-rank-tests}).

\begin{table*}[!htbp]
\centering
\caption{Cross-dataset Friedman tests. $Q$ is the tie-corrected Friedman statistic.}
\label{tab:global-rank-tests}
\small
\begin{tabular}{@{}lrrr@{}}
\toprule
Evaluation measure & Datasets & $Q$ & $p$-value \\
\midrule
TDC leaderboard & 22 & 107.13 & $2.02\times10^{-18}$ \\
Calibrated Brier & 13 & 45.75 & $1.59\times10^{-6}$ \\
Calibrated interval score & 9 & 54.40 & $4.08\times10^{-8}$ \\
\bottomrule
\end{tabular}
\end{table*}

Table~\ref{tab:global-rank-holm} gives every mean rank and the Holm-adjusted comparison with ensemble
MC dropout. The adjustment includes all 55 pairwise comparisons within each evaluation measure.

\begin{table*}[!htbp]
\centering
\caption{Mean ranks and Holm-adjusted comparisons with ensemble MC dropout. Bold adjusted
$p$-values indicate a significant difference at the 5\% level. The adjustment accounts for all 55
pairwise method comparisons within each evaluation measure.}
\label{tab:global-rank-holm}
\scriptsize
\setlength{\tabcolsep}{4pt}
\begin{tabular}{@{}lrrrrrr@{}}
\toprule
& \multicolumn{2}{c}{TDC leaderboard} & \multicolumn{2}{c}{Brier} & \multicolumn{2}{c}{Interval score} \\
\cmidrule(lr){2-3}\cmidrule(lr){4-5}\cmidrule(l){6-7}
Method & Rank & $p_{\mathrm{Holm}}$ & Rank & $p_{\mathrm{Holm}}$ & Rank & $p_{\mathrm{Holm}}$ \\
\midrule
LightGBM & 6.36 & \textbf{0.0097} & 4.23 & 1.0000 & 3.56 & 1.0000 \\
Deterministic MLP & 8.70 & \textbf{$<0.0001$} & 8.38 & \textbf{0.0101} & 10.33 & \textbf{$<0.0001$} \\
MC dropout & 7.91 & \textbf{$<0.0001$} & 7.69 & 0.0690 & 5.67 & 1.0000 \\
SWAG & 7.32 & \textbf{0.0002} & 7.38 & 0.1431 & 7.78 & \textbf{0.0486} \\
Laplace & 8.16 & \textbf{$<0.0001$} & 9.23 & \textbf{0.0007} & 9.22 & \textbf{0.0014} \\
Post-hoc VBLL & 8.82 & \textbf{$<0.0001$} & 6.46 & 0.9364 & 8.33 & \textbf{0.0142} \\
Deep ensemble & 4.20 & 1.0000 & 4.54 & 1.0000 & 6.00 & 1.0000 \\
Ensemble MC dropout & 2.73 & -- & 3.54 & -- & 2.67 & -- \\
MultiSWAG & 3.86 & 1.0000 & 5.46 & 1.0000 & 5.00 & 1.0000 \\
MoLA & 3.20 & 1.0000 & 5.46 & 1.0000 & 3.44 & 1.0000 \\
Ensemble post-hoc VBLL & 4.73 & 1.0000 & 3.62 & 1.0000 & 4.00 & 1.0000 \\
\bottomrule
\end{tabular}
\end{table*}

\subsection{Computational scale}

Let $U=22\times10\times5\times5=5{,}500$ denote the number of
dataset--partition--model-seed--outer-fold units. Table~\ref{tab:compute-counts} counts shared
neural sweeps and ensemble backbones once, irrespective of how many UQ readouts they produce.

\begin{table}[!htbp]
\centering
\caption{Computation underlying the eleven-pipeline comparison.}
\label{tab:compute-counts}
\small
\begin{tabularx}{\linewidth}{@{}Xr@{}}
\toprule
Computation & Count \\
\midrule
LightGBM inner-CV fits & 880,000 \\
LightGBM outer refits & 5,500 \\
Shared MLP inner-CV fits & 880,000 \\
Single-model MLP outer refits & 5,500 \\
Ensemble-member MLP refits & 55,000 \\
\textbf{From-scratch model fits} & \textbf{1,826,000} \\
Checkpoint-resumed SWAG continuations & 60,500 \\
Post-hoc VBLL last-layer grid-point fits & 4,900,500 \\
Calibration applications & 60,500 \\
Outer-fold method evaluations & 60,500 \\
Score evaluations across the three main measures & 121,000 \\
\bottomrule
\end{tabularx}
\end{table}

The 55,000 ensemble-member entries are full MLP refits. The larger post-hoc VBLL count consists of
81 last-layer optimizations for each of 60,500 frozen feature maps, comprising 5,500 single models
and 55,000 ensemble members. Each optimization fits one prior-scale and Wishart-scale pair without
retraining the MLP backbone.
\FloatBarrier

\subsection{Additional evaluation measures}

\begin{figure*}[p]
  \centering
  \includegraphics[width=\textwidth,height=0.82\textheight,keepaspectratio]{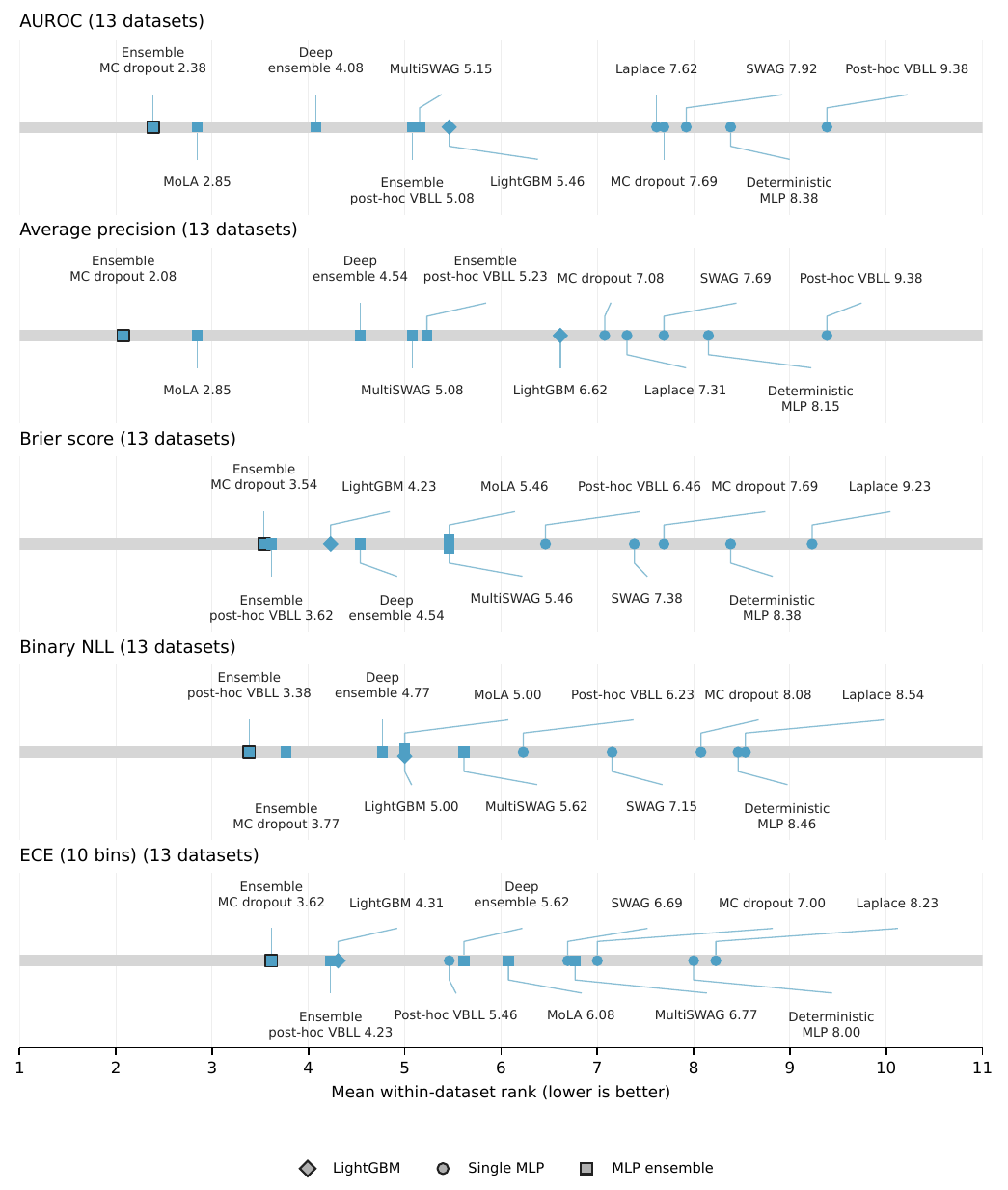}
  \caption{Mean ranks for five binary evaluation measures after temperature scaling. Each measure
  contains the same 13 datasets and eleven methods. The outlined marker identifies the lowest
  observed mean rank; coincident points have equal mean ranks.}
  \label{fig:binary-metric-rails}
\end{figure*}

\begin{figure*}[p]
  \centering
  \includegraphics[width=\textwidth]{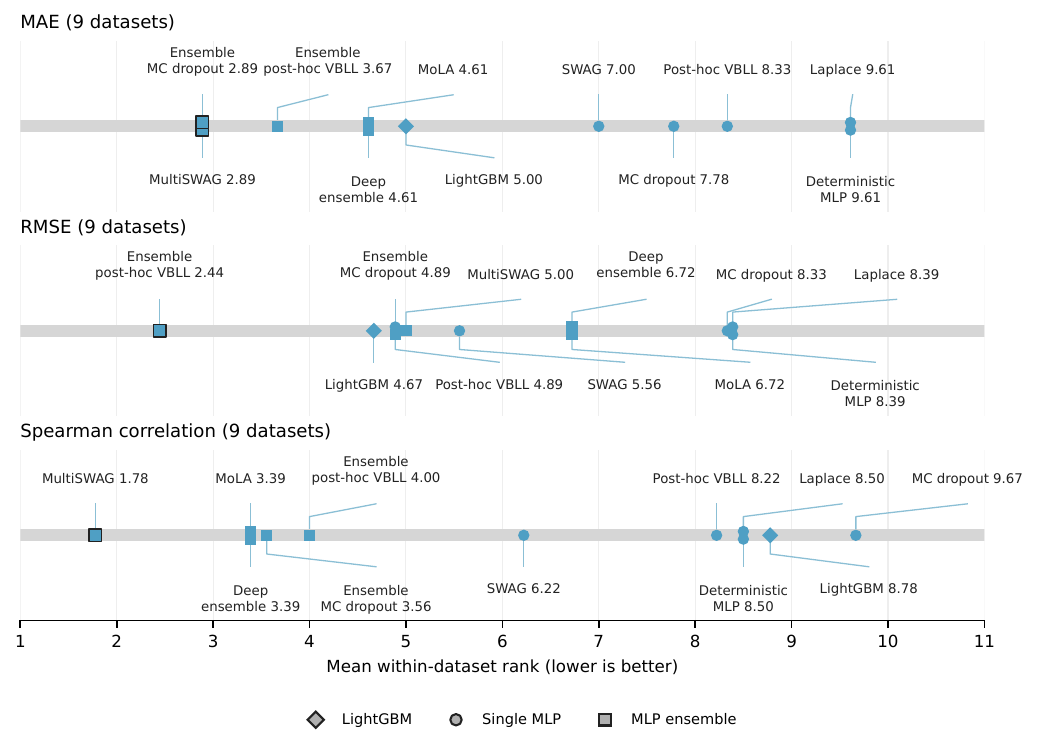}
  \caption{Mean ranks for regression point-prediction measures. Each measure contains the same nine
  datasets and eleven methods. The outlined marker identifies the lowest observed mean rank;
  coincident points have equal mean ranks.}
  \label{fig:regression-point-metric-rails}
\end{figure*}

\begin{figure*}[p]
  \centering
  \includegraphics[width=\textwidth]{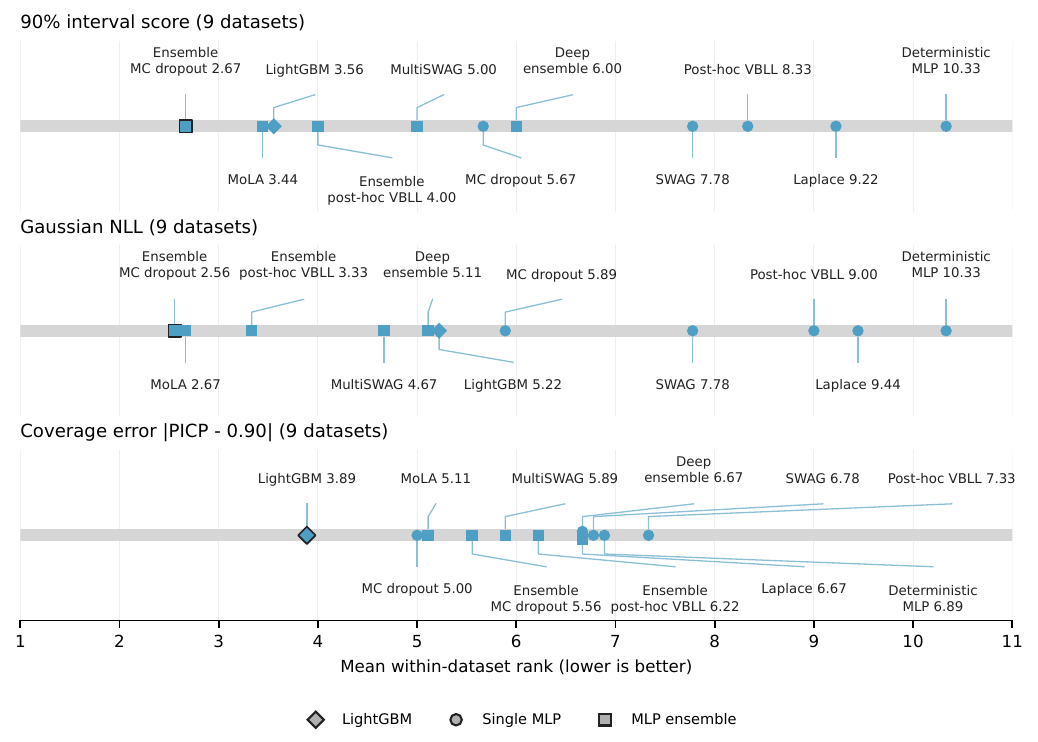}
  \caption{Mean ranks for interval score, Gaussian NLL, and absolute coverage error after
  method-specific regression calibration. Each measure contains the same nine datasets and eleven
  methods. The outlined marker identifies the lowest observed mean rank; coincident points have
  equal mean ranks.}
  \label{fig:regression-uq-metric-rails}
\end{figure*}

\begin{figure*}[!htbp]
  \centering
  \includegraphics[width=\textwidth,height=.76\textheight,keepaspectratio]{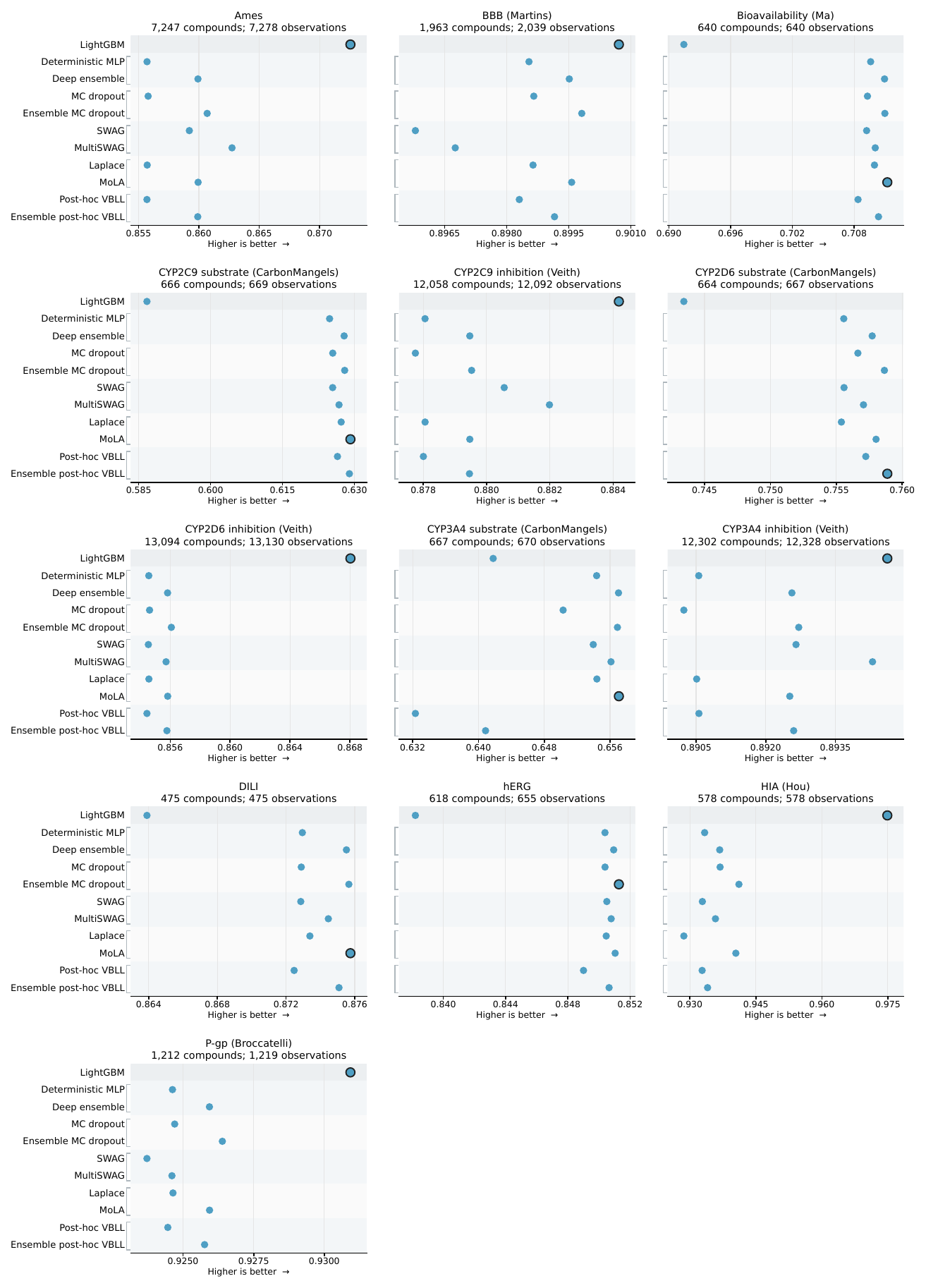}
  \caption{Temperature-calibrated area under the receiver operating characteristic curve (AUROC)
  on all 13 binary datasets. Higher is better. Each marker is the mean of 250 outer-fold scores.
  Outlined markers identify empirical bests.}
  \label{fig:appendix-native-binary-auroc}
\end{figure*}

\begin{figure*}[p]
  \centering
  \includegraphics[width=\textwidth,height=.88\textheight,keepaspectratio]{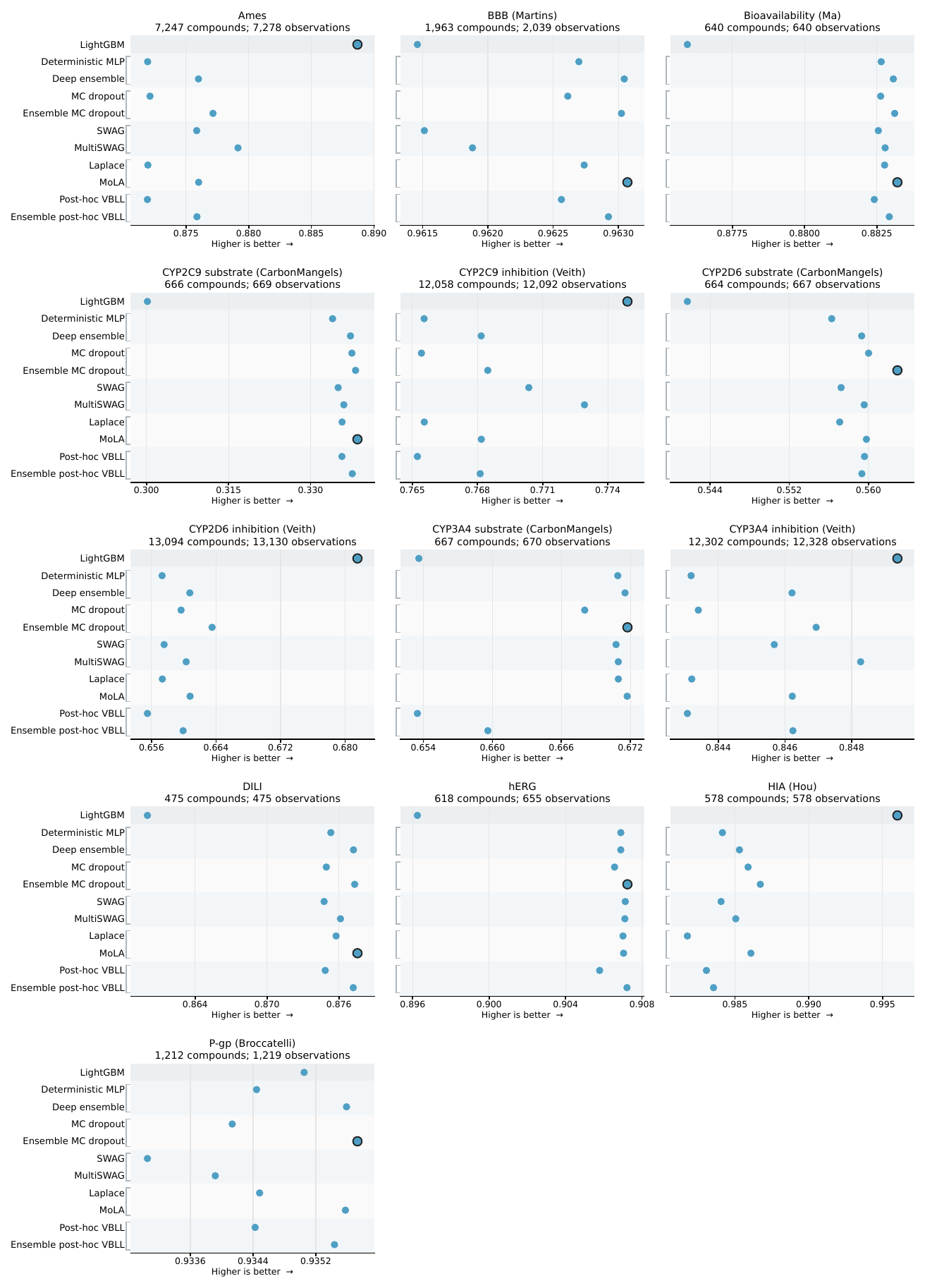}
  \caption{Temperature-calibrated average precision on all 13 binary datasets. Higher is better.
  Each marker is the mean of 250 outer-fold scores. Outlined markers identify empirical bests.}
  \label{fig:appendix-native-binary-ap}
\end{figure*}

\begin{figure*}[p]
  \centering
  \includegraphics[width=\textwidth,height=.88\textheight,keepaspectratio]{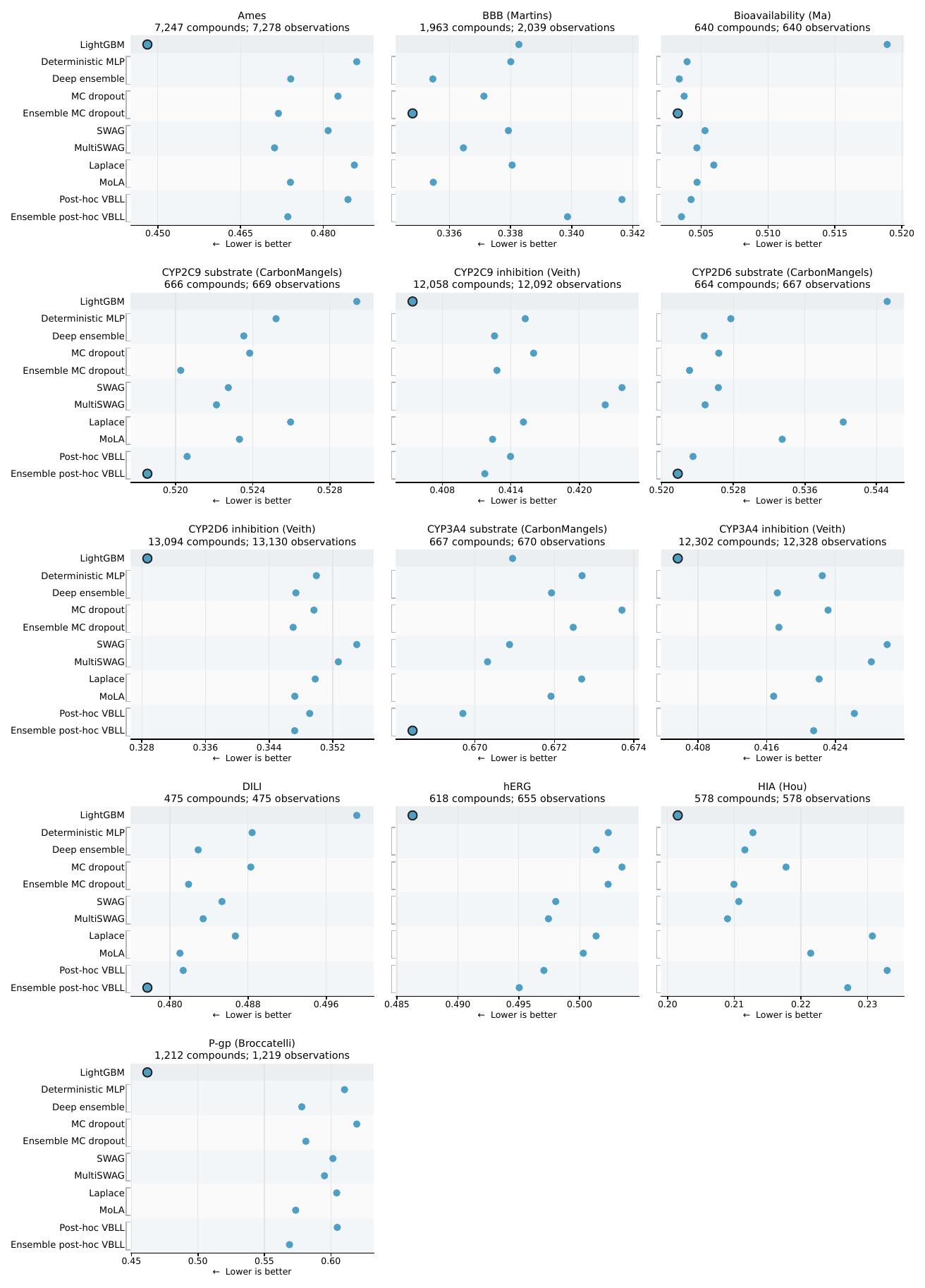}
  \caption{Temperature-calibrated binary negative log-likelihood on all 13 binary datasets.
  Lower is better. Each marker is the mean of 250 outer-fold scores. Outlined markers identify
  empirical bests.}
  \label{fig:appendix-native-binary-nll}
\end{figure*}

\begin{figure*}[p]
  \centering
  \includegraphics[width=\textwidth,height=.88\textheight,keepaspectratio]{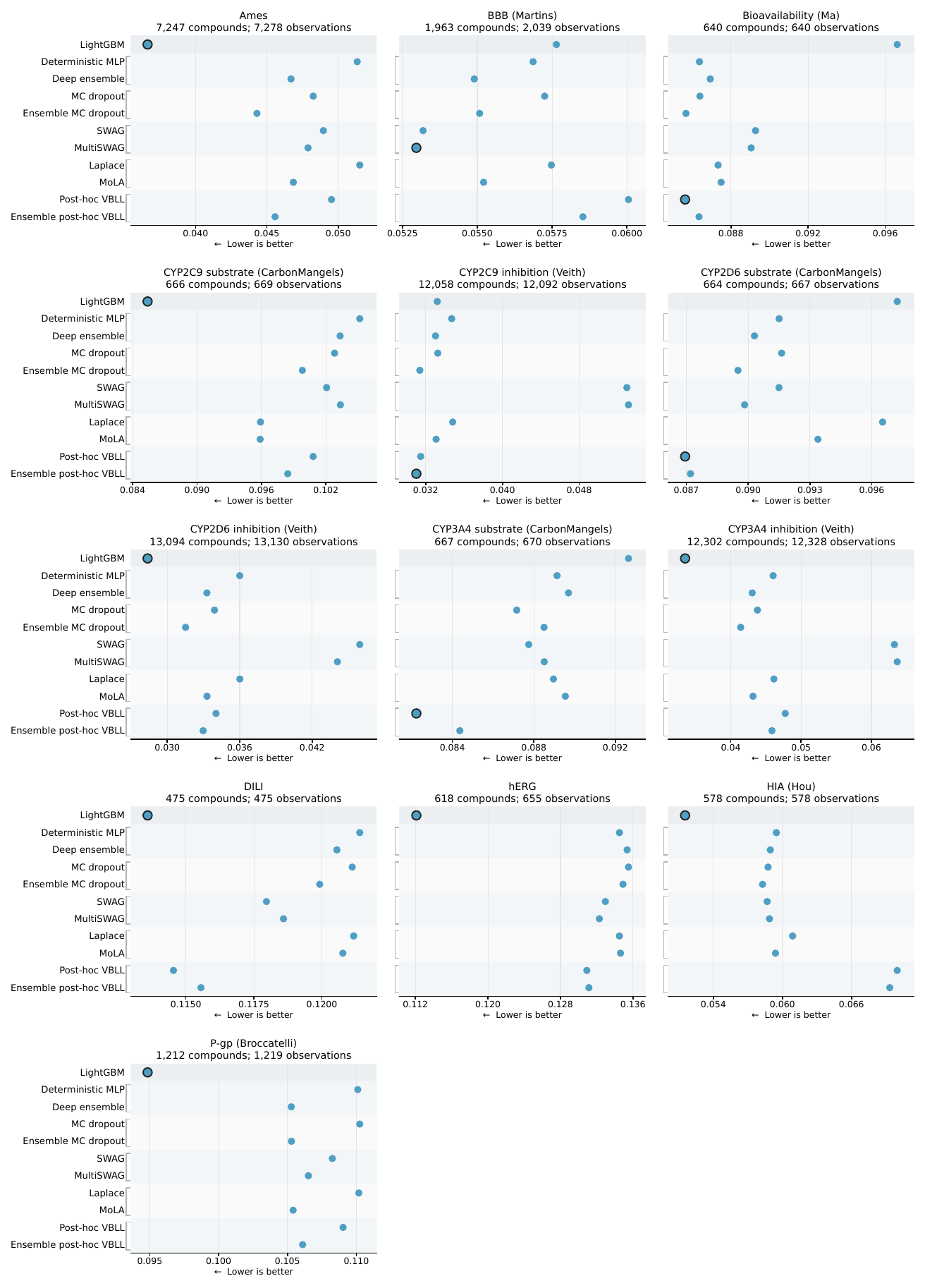}
  \caption{Temperature-calibrated 10-bin expected calibration error on all 13 binary datasets.
  Lower is better. Each marker is the mean of 250 outer-fold scores. Outlined markers identify
  empirical bests.}
  \label{fig:appendix-native-binary-ece}
\end{figure*}

\begin{figure*}[p]
  \centering
  \includegraphics[width=\textwidth,height=.88\textheight,keepaspectratio]{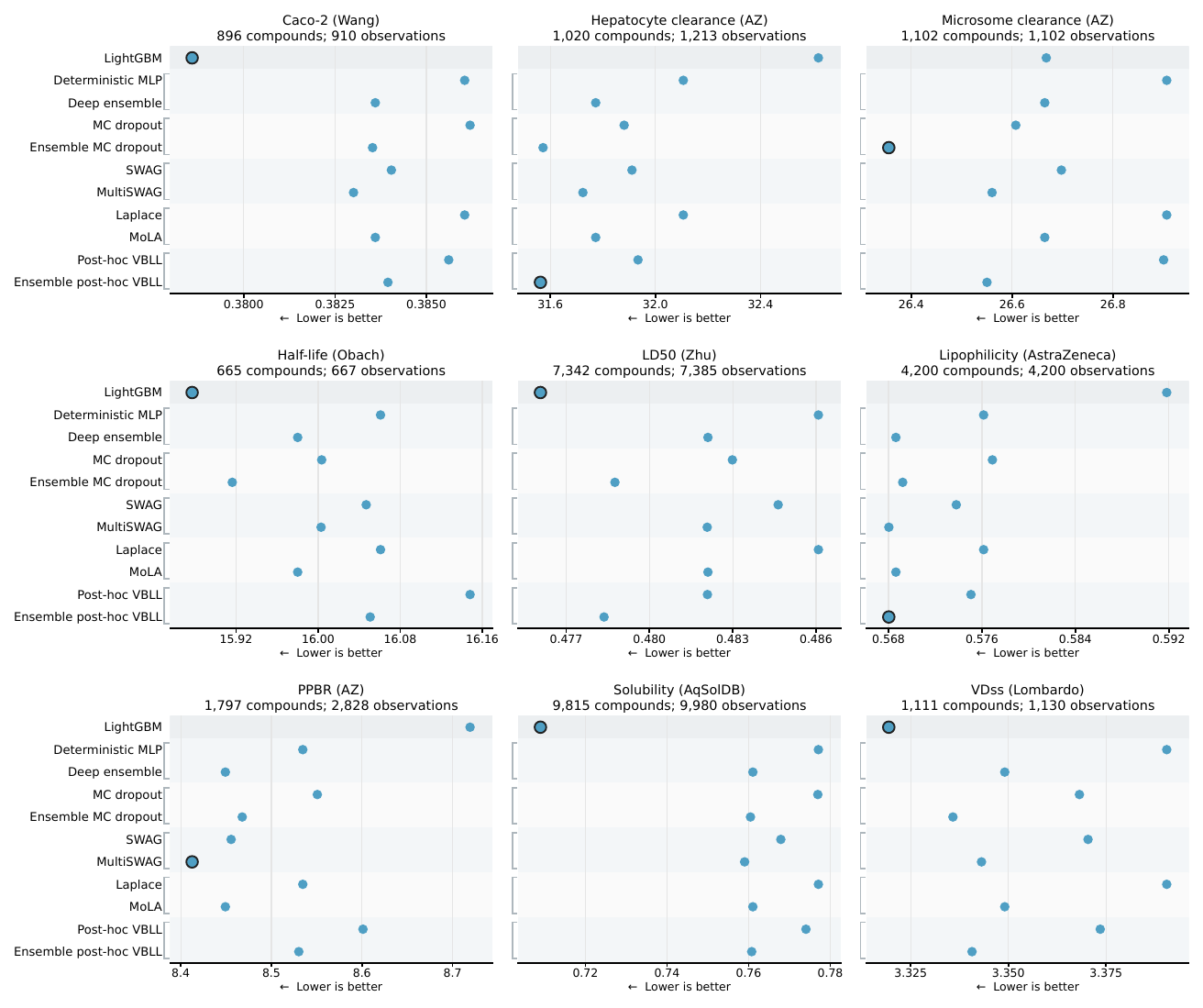}
  \caption{Mean absolute error on all nine regression datasets under the fixed method-specific
  calibration pipelines. Lower is better. Each marker is the mean of 250 outer-fold scores. The
  outlined markers identify empirical bests.}
  \label{fig:appendix-native-regression-mae}
\end{figure*}

\begin{figure*}[p]
  \centering
  \includegraphics[width=\textwidth,height=.88\textheight,keepaspectratio]{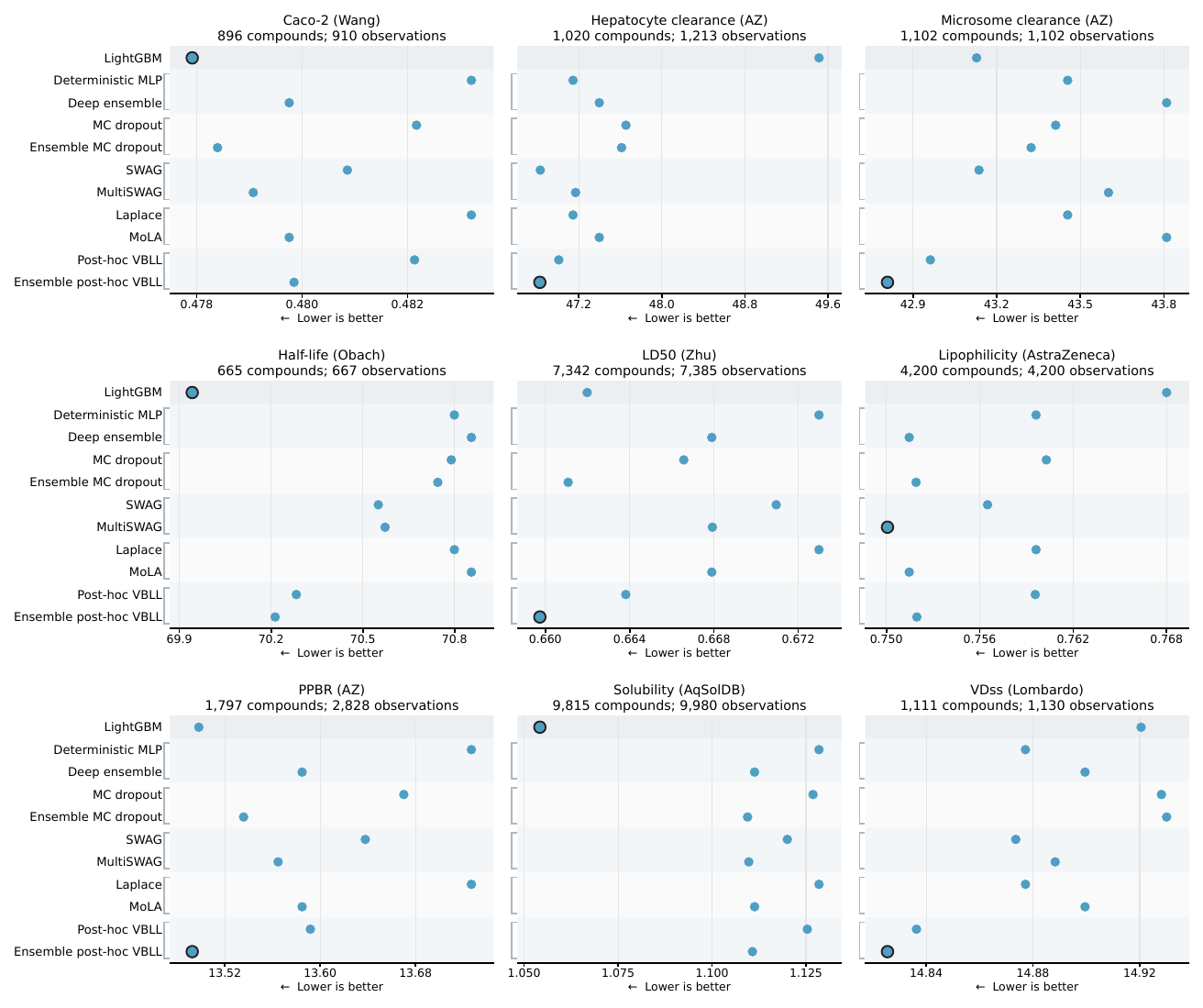}
  \caption{Root mean squared error on all nine regression datasets under the fixed
  method-specific calibration pipelines. Lower is better. Each marker is the mean of 250
  outer-fold scores. Outlined markers identify empirical bests.}
  \label{fig:appendix-native-regression-rmse}
\end{figure*}

\begin{figure*}[p]
  \centering
  \includegraphics[width=\textwidth,height=.88\textheight,keepaspectratio]{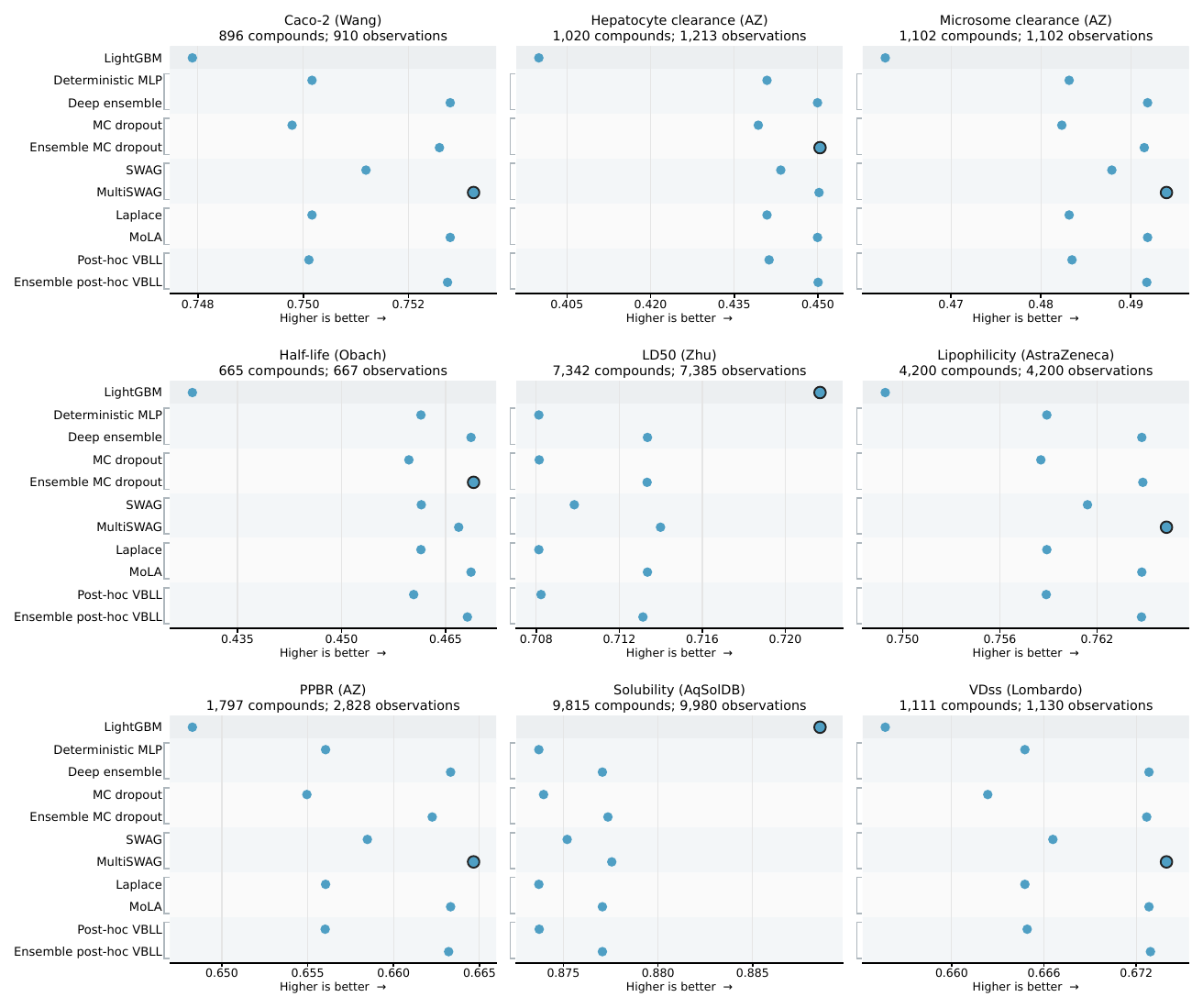}
  \caption{Spearman correlation on all nine regression datasets under the fixed method-specific
  calibration pipelines. Higher is better. Each marker is the mean of 250 outer-fold scores. The
  outlined markers identify empirical bests.}
  \label{fig:appendix-native-regression-spearman}
\end{figure*}

\begin{figure*}[p]
  \centering
  \includegraphics[width=\textwidth,height=.88\textheight,keepaspectratio]{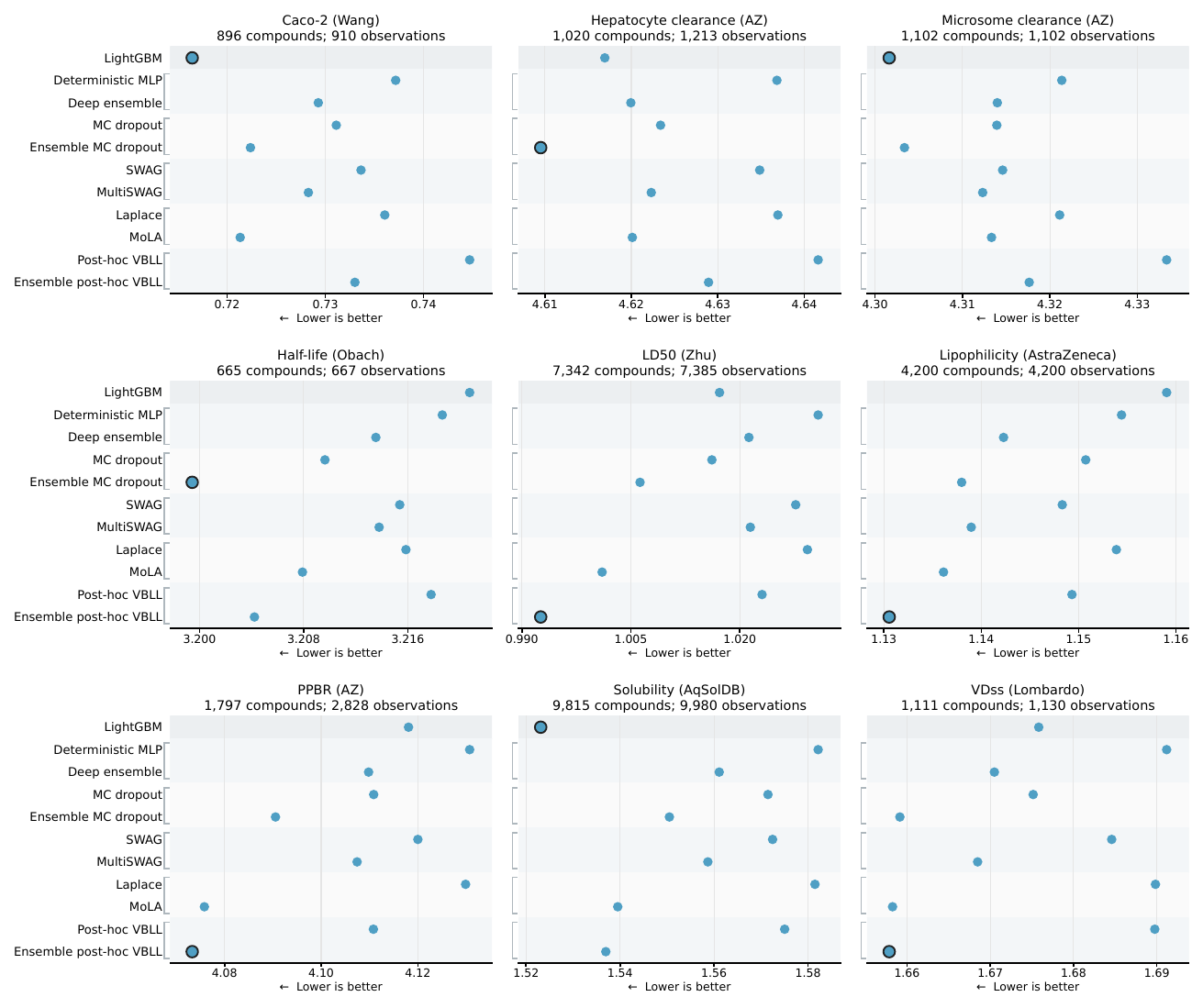}
  \caption{Gaussian negative log-likelihood on all nine regression datasets under the fixed
  method-specific calibration map. Lower is better. Each marker is the mean of 250 outer-fold
  scores. Outlined markers identify empirical bests.}
  \label{fig:appendix-native-regression-nll}
\end{figure*}

\begin{figure*}[p]
  \centering
  \includegraphics[width=\textwidth,height=.88\textheight,keepaspectratio]{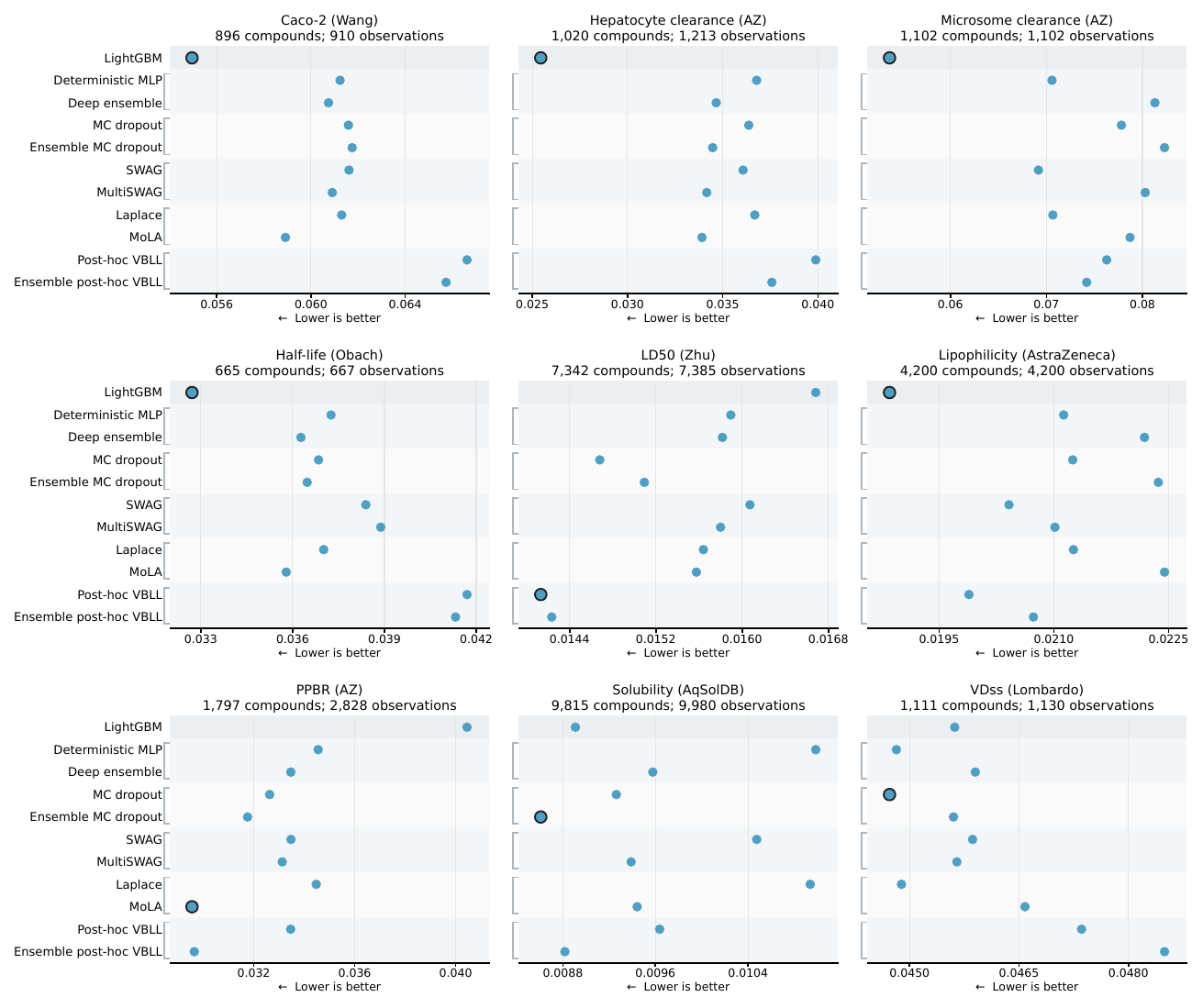}
  \caption{Absolute 90\% prediction interval coverage probability (PICP) error,
  $|\mathrm{PICP}-0.90|$, on all nine regression datasets under the fixed method-specific
  calibration map. Lower is better. Each marker is the
  mean of 250 outer-fold scores. Outlined markers identify empirical bests.}
  \label{fig:appendix-native-regression-coverage-error}
\end{figure*}

Figures~\ref{fig:binary-metric-rails}--\ref{fig:regression-uq-metric-rails} show mean ranks for the
additional measures listed in Section~\ref{sec:methods}. Each score averages the same 250 outer-fold
values as the main analysis. The corresponding values on each dataset's original scale appear in
Figures~\ref{fig:appendix-native-binary-auroc}--\ref{fig:appendix-native-regression-coverage-error}.

\FloatBarrier

\section{Hyperparameter optimization procedure}
\label{app:hpo-adequacy}

\paragraph{Search allocation.}
LightGBM and the deterministic MLP are selected from 32 configurations generated by a scrambled
Sobol sequence with seed 0. Each configuration is evaluated on all five inner folds, and selection
uses the mean inner-fold TDC metric and the fold-averaged validation curve. MC dropout, SWAG, and
Laplace inherit the deterministic MLP selection. For post-hoc VBLL, the nine prior scales are
logarithmically spaced from $10^{-2}$ to $10^2$ and the nine Wishart scales from $10^{-4}$ to $1$.
Each of the 81 pairs is fitted for 1,000 full-batch Adam steps on the model-development rows, and the pair
with the largest final ELBO is used. Each ensemble inherits the selection of its corresponding
single-model pipeline.

\begin{table}[!htbp]
\centering
\caption{Hyperparameter search spaces for LightGBM and the deterministic MLP. Uniform and
log-uniform dimensions are mapped from the Sobol points; discrete levels receive equal intervals of
the unit range.}
\label{tab:hpo-spaces}
\small
\begin{tabularx}{\textwidth}{@{}lllX@{}}
\toprule
Model & Hyperparameter & Sampling & Range or levels \tabularnewline
\midrule
LightGBM & \texttt{learning\_rate} & Log-uniform & $[10^{-2},\,0.3]$ \tabularnewline
LightGBM & \texttt{num\_leaves} & Log-uniform integer & $[15,\,255]$ \tabularnewline
LightGBM & \texttt{min\_child\_samples} & Uniform integer & $[5,\,100]$ \tabularnewline
LightGBM & \texttt{feature\_fraction} & Uniform & $[0.4,\,1.0]$ \tabularnewline
LightGBM & \texttt{bagging\_fraction} & Uniform & $[0.5,\,1.0]$ \tabularnewline
LightGBM & \texttt{lambda\_l1} & Log-uniform & $[10^{-3},\,10]$ \tabularnewline
LightGBM & \texttt{lambda\_l2} & Log-uniform & $[10^{-3},\,10]$ \tabularnewline
\addlinespace
MLP & \texttt{n\_layers} & Discrete & $\{1,2,3,4\}$ \tabularnewline
MLP & \texttt{hidden\_dim} & Powers of two & $\{128,256,512,1024\}$ \tabularnewline
MLP & \texttt{dropout} & Uniform & $[0.10,\,0.75]$ \tabularnewline
MLP & \texttt{learning\_rate} & Log-uniform & $[10^{-5},\,10^{-3}]$ \tabularnewline
MLP & \texttt{weight\_decay} & Log-uniform & $[10^{-6},\,10^{-2}]$ \tabularnewline
MLP & \texttt{batch\_size} & Powers of two & $\{16,32,64,128,256\}$ \tabularnewline
\bottomrule
\end{tabularx}
\end{table}
\FloatBarrier

\paragraph{Stability of inner-fold selection.}
Using the stored inner-validation trajectories, we recalculated model selection for every LightGBM
and deterministic-MLP outer-fold fit in the $10\times5$ grid. We compared the hyperparameters and
training iteration selected from one or four inner folds with those selected from all five.
Table~\ref{tab:inner-selection-stability} reports equal-dataset medians over all 22 datasets. A
single fold selects the same configuration as all five folds in 19.2\% of LightGBM cases and 17.8\%
of MLP cases. Using four folds raises these rates to 67.2\% and 59.3\%. Across the five individual
folds, the median outer evaluation contains four distinct configurations and five distinct
configuration--iteration pairs for both models. Raw differences between stopping iterations are
not reported because most folds also choose a different configuration and therefore a different
training trajectory.

\begin{table}[!htbp]
\centering
\caption{Stability of inner-CV selection relative to the choice made from all five folds. Values
are medians of the 22 dataset-level summaries. Joint agreement requires both the configuration and
stopping iteration to match. Relative loss divides the five-fold validation-objective loss by the
gap between the best and median configuration within the same sweep.}
\label{tab:inner-selection-stability}
\scriptsize
\setlength{\tabcolsep}{3pt}
\begin{tabular}{@{}lrrrrrr@{}}
\toprule
& \multicolumn{3}{c}{One inner fold} & \multicolumn{3}{c}{Four inner folds} \\
\cmidrule(lr){2-4}\cmidrule(l){5-7}
Model & Config. agree & Joint agree & Relative loss & Config. agree & Joint agree & Relative loss \\
\midrule
LightGBM & 19.2\% & 1.0\% & 0.911 & 67.2\% & 24.3\% & 0.027 \\
MLP & 17.8\% & 2.6\% & 1.281 & 59.3\% & 36.0\% & 0.056 \\
\bottomrule
\end{tabular}
\end{table}

Subsequent refitting used the configuration selected by all five folds. The table therefore measures
how often reduced inner-CV designs reproduce that selection. The stored results do not compare the
outer-test performance of their alternative choices.

\section{Scores by dataset}
\label{app:dataset-scores}

The figures show every dataset and all eleven methods on the original metric scale. Each score
marker is the mean of 250 outer-fold scores. For MCB, we average the five model seeds and five
outer folds within each chemical partition, giving 10 paired complete-partition scores. For
residual mean square (MSE), the one-step Hsu decision threshold is
$H=q_{1,11}\sqrt{\mathrm{MSE}/10}$, where $q_{1,11}$ is the corresponding 95\% MCB critical
value for eleven methods with 90 residual degrees of freedom. The pale region starts at the empirical winner and
extends by $H$ toward worse scores. A method inside the region cannot be ruled out as best; a method
beyond its boundary is classified as significantly worse by Hsu's test.

\begin{figure}[!htbp]
  \centering
  \includegraphics[width=\linewidth,height=.86\textheight,keepaspectratio]{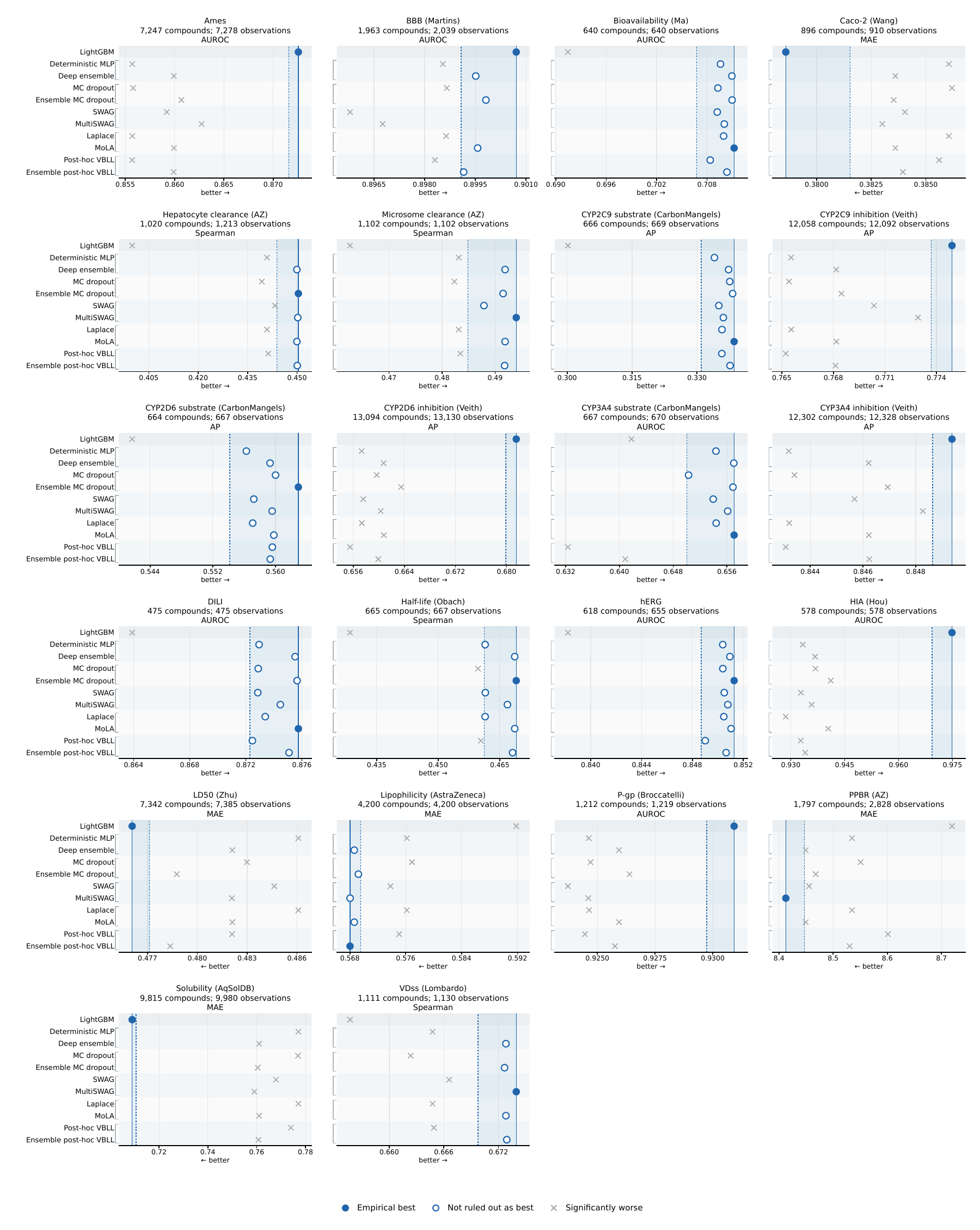}
  \caption{Full-grid TDC leaderboard metric values for all 22 ADMET datasets. The arrow below each
  heading gives the metric direction. Filled circles mark the empirical best, open circles mark other
  methods not ruled out as best, and crosses mark methods classified as significantly worse. The
  pale region shows the MCB decision threshold described above.}
  \label{fig:tdc-dataset-scores}
\end{figure}

\begin{figure}[!htbp]
  \centering
  \includegraphics[width=\linewidth,height=.86\textheight,keepaspectratio]{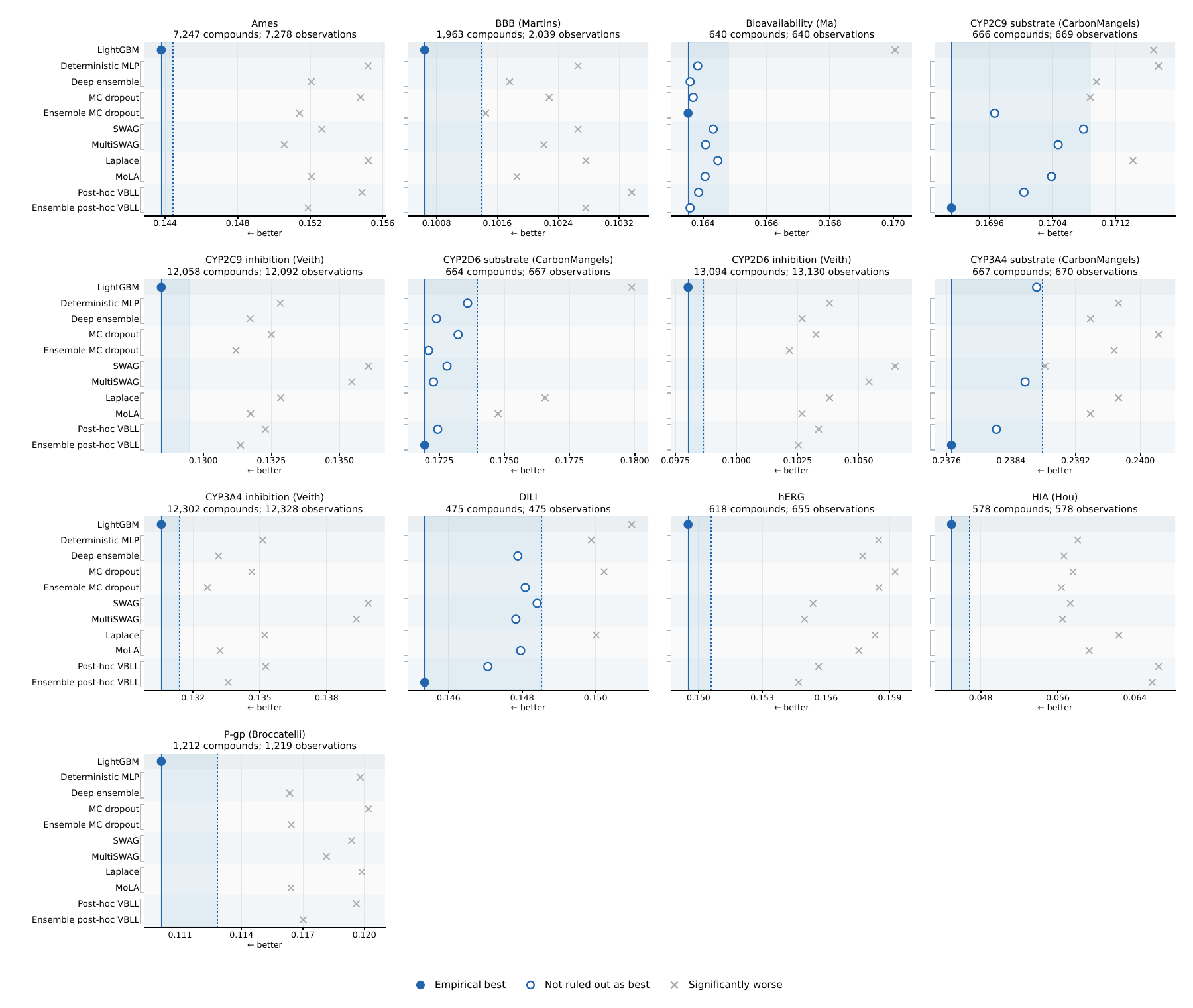}
  \caption{Full-grid temperature-calibrated Brier scores for all 13 binary ADMET datasets. Lower is
  better. Marks and the decision region are defined as in Figure~\ref{fig:tdc-dataset-scores}. A common
  scalar temperature is fitted on each dedicated calibration holdout.}
  \label{fig:binary-dataset-scores}
\end{figure}

\clearpage
\begin{figure}[!htbp]
  \centering
  \includegraphics[width=\linewidth,height=.86\textheight,keepaspectratio]{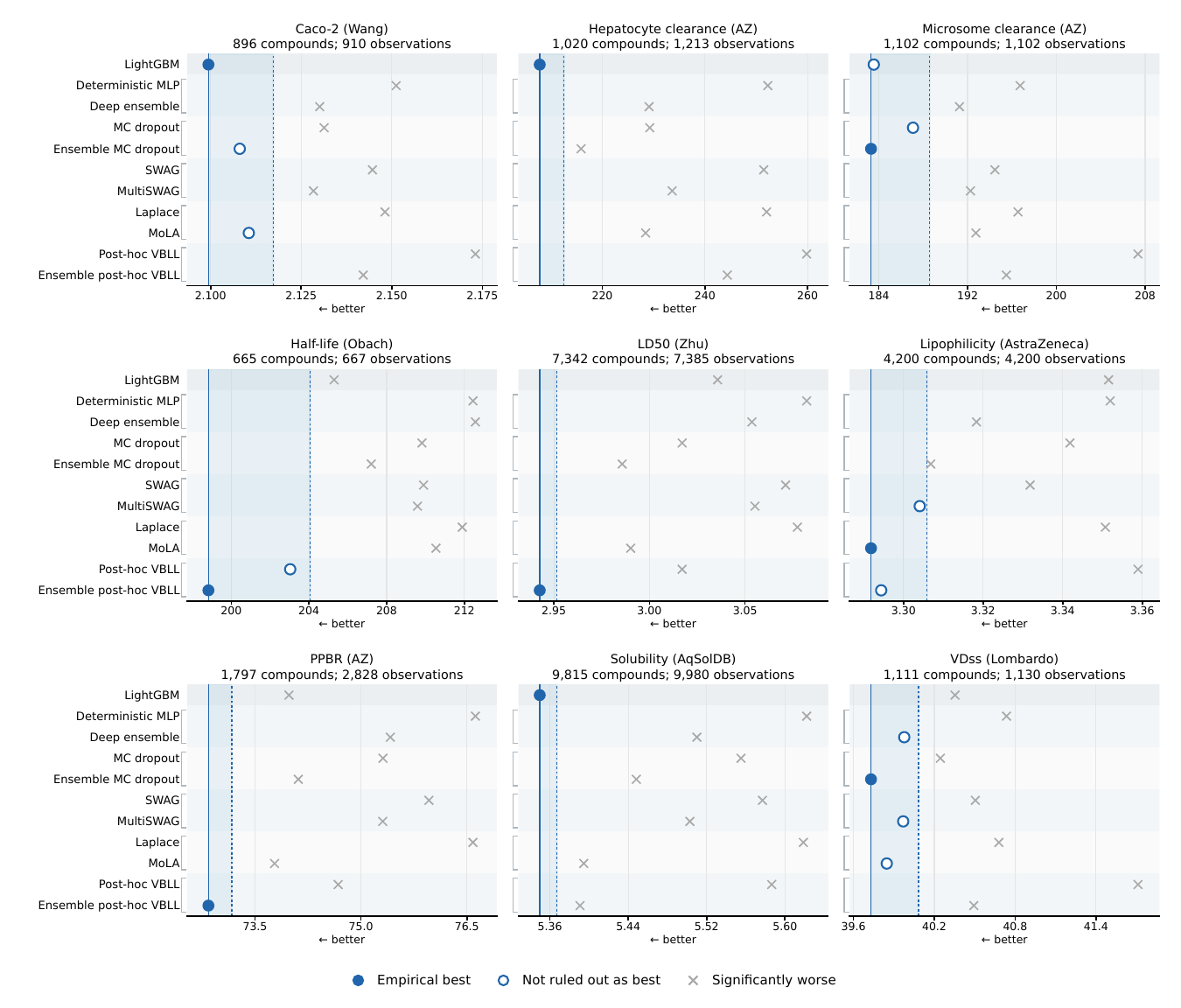}
  \caption{Full-grid method-specifically calibrated 90\% interval scores for all nine regression
  ADMET datasets. Lower is better. Marks and the decision region are defined as in
  Figure~\ref{fig:tdc-dataset-scores}. The horizontal axes differ because interval score is reported
  in the original units of each endpoint.}
  \label{fig:regression-dataset-scores}
\end{figure}

\FloatBarrier

Under the assumed block model, Hsu's selection rule includes at least one method with the best
population mean score with probability at least 95\% \citep[p.~1141]{hsu1984constrained}.
Excluding any particular method is a 5\%-level test of whether it performs worse than the best
alternative. If several methods share the optimal population mean, the 95\% guarantee concerns
including a best method, not simultaneously including all of them.

\subsection{Additional metrics on their original scales}

The following figures complement the mean-rank rails with the underlying full-grid metric values.
They use the same fixed method order and pairing as the three formal score sheets. Outlined
markers identify empirical bests. Horizontal axes are fitted separately for each dataset because
several regression metrics have
different endpoint units and scales.

\FloatBarrier

\end{document}